\documentclass[lettersize,journal]{IEEEtran}

\usepackage{enumitem}
\usepackage{longtable}
\usepackage{booktabs}
\usepackage{threeparttable}
\usepackage{adjustbox}
\usepackage{tabularx}
\usepackage{multicol}

\usepackage{orcidlink}
\usepackage{amsmath,amsfonts}
\usepackage{algorithmic}
\usepackage{algorithm}
\usepackage{array}
\usepackage[caption=false,font=footnotesize]{subfig}
\usepackage{textcomp}
\usepackage{stfloats}
\usepackage{verbatim}
\usepackage{graphicx}
\usepackage{cite}
\begin{document}

\title{Deep Time‑Series Models Meet Volatility: Multi‑Horizon Electricity Price Forecasting in the Australian National Electricity Market}

\author{Mohammed Osman Gani\,\orcidlink{0000-0002-9918-8589}, Zhipeng He\,\orcidlink{0000-0002-8241-8499}, Chun Ouyang\,\orcidlink{0000-0001-7098-5480}, and Sara Khalifa\,\orcidlink{0000-0002-3417-2834}
\thanks{The authors are with the School of Information Systems and the Energy Transition Centre, Queensland University of Technology, Brisbane, QLD 4000, Australia (e-mail: mohammedosman.gani@hdr.qut.edu.au; zippo.he@qut.edu.au; c.ouyang@qut.edu.au; sara.khalifa@qut.edu.au).}
\thanks{Manuscript received Month xx, 202x; revised Month xx, 202x.}%
}

\markboth{~Vol.~xx, No.~x, Month~202x}%
{Gani \MakeLowercase{\textit{et al.}}:Deep Time‑Series Models Meet Volatility}


\maketitle

\begin{abstract}
Accurate electricity price forecasting~(EPF) is increasingly difficult in markets characterised by extreme volatility, frequent price spikes, and rapid structural shifts. Deep learning (DL) has been increasingly adopted in EPF due to its ability to achieve high forecasting accuracy. Recently, state-of-the-art (SOTA) deep time-series models have demonstrated promising performance across general forecasting tasks. Yet, their effectiveness in highly volatile electricity markets remains underexplored. Moreover, existing EPF studies rarely assess how model accuracy varies across intraday periods, leaving model sensitivity to market conditions unexplored. 
To address these gaps, this paper proposes an EPF framework that systematically evaluates SOTA deep time‑series models using a direct multi‑horizon forecasting approach across day‑ahead and two‑day‑ahead settings. We conduct a comprehensive empirical study across all five regions of the Australian National Electricity Market using contemporary, high-volatility data.
The results reveal a clear gap between time-series benchmark expectations and observed performance under real-world price volatility: recent deep time‑series models often fail to surpass standard DL baselines. All models experience substantial degradation under extreme and negative prices, yet DL baselines often remain competitive. Intraday performance analysis further reveals that all evaluated models are consistently vulnerable to prevailing market conditions, where absolute errors peak during evening ramps, relative errors escalate during midday negative‑price periods, and directional accuracy deteriorates sharply during abrupt shifts in price direction. These findings emphasise the need for volatility‑aware modelling strategies and richer feature representations to advance EPF. 
\end{abstract}


\section{Introduction}
Electricity price forecasting (EPF) is fundamental to modern power systems, supporting optimised trading strategies, asset scheduling, risk management, and system reliability. Yet accurate EPF remains highly challenging because electricity prices exhibit pronounced volatility, abrupt spikes, and frequent regime shifts, arising from intricate interactions among demand–supply conditions and market design~\cite{Lago2018-tc}. These challenges have intensified in recent years, driven by the rapid expansion of variable renewable energy, disruptions during and after COVID‑19, and global political and fuel‑market instability, all of which have collectively contributed to greater uncertainty and increased the frequency of extreme price movements~\cite{Numminen2025-wf}. As a result, electricity prices can fluctuate from sharply negative levels to upward spikes of several thousand dollars per megawatt‑hour, producing extremely volatile and structurally complex time series, making electricity price modelling exceptionally difficult.

As electricity markets grow increasingly complex, deep learning (DL) models~\cite{Chai2024-gi} have emerged as the dominant modelling choice in EPF, surpassing statistical and traditional machine learning approaches due to their ability to capture strong nonlinearities and support long‑term forecasting. In parallel, the time‑series forecasting field has seen rapid advances in model architectures capable of long‑sequence learning and efficient temporal representation, with many achieving state‑of‑the‑art (SOTA) performance on widely used benchmark datasets~\cite{Wang2024-ky}. However, these benchmark datasets are typically far more stable and structured than electricity price series and do not exhibit the extreme volatility, negative prices, or sudden regime shifts, which are typical characteristics of electricity prices. Consequently, it remains an open question whether these recent advances in deep time‑series modelling can generalise to such highly volatile and structurally complex time series. This aspect has been largely overlooked in existing EPF research, where deep time‑series models have not been evaluated under the extreme, rapidly evolving conditions observed in electricity markets. 

Moreover, most recent work~\cite{Li2022-nq, Heidarpanah2023-ax, Pourdaryaei2024-kb, Li2025-zw, Perla2025-he} reports model performance by averaging errors across forecast steps. Because of this aggregated evaluation, it remains unexplored whether forecasting performance is uniform throughout the day or whether models are affected by underlying market conditions, particularly in recent circumstances characterised by abrupt regime shifts. Therefore, it is essential to evaluate models at a granular intraday interval level to diagnose the possibility of time‑of‑day forecasting variation and sensitivity to market conditions. Such an evaluation is critical for assessing operational reliability and identifying model strengths and weaknesses to support future methodological advancements. This will enable more targeted model improvements and better alignment with real‑world decision‑making needs. 

To ensure that such intraday diagnostics are reliable, the evaluation must be conducted on datasets that reflect current market dynamics. However, many recent EPF studies evaluate models using historical data that no longer capture present market conditions. This is especially evident in recent research on the Australian National Electricity Market (NEM)~\cite{Tan2023-ck, Ghimire2024-ou, Li2025-zw, Perla2025-he}, where studies frequently rely on pre‑2022 data—before major NEM policy changes, post‑COVID demand shifts, and the sharp rise in renewable penetration. Since these developments have dramatically increased market volatility and transformed underlying price dynamics, it is essential to evaluate models under current market conditions to better understand their behaviour and robustness.

This paper addresses the above gaps by proposing a DL‑based EPF framework that systematically examines the effectiveness of state‑of‑the‑art deep time‑series models under contemporary market conditions, using data from all five NEM regions. The framework compares recent SOTA time‑series architectures from the forecasting literature with standard DL models commonly adopted in EPF, enabling a rigorous assessment of whether advances in time‑series modelling translate to highly volatile electricity markets. We evaluate model performance in both the conventional day‑ahead setting and an extended two‑day‑ahead horizon, thereby assessing the capability for longer‑term forecasting. Throughout the study, we adopt a direct multi‑horizon forecasting approach. Alongside standard evaluation and the analysis of extreme and negative prices, we further assess performance separately for each of the 48 half‑hour intervals of the day, enabling detailed intraday analysis that reveals time‑of‑day‑specific forecasting challenges and reveals the models’ sensitivity to prevailing market conditions. 

Based on the above, this study makes several key contributions to the EPF literature, summarised as follows.
\begin{enumerate}[label=\textbullet]
    \item We evaluate the effectiveness of SOTA deep time‑series models under contemporary high‑volatility electricity market conditions. 
    \item We examine how forecasting performance varies with underlying market conditions by evaluating model accuracy across intraday intervals, enabling detailed insights into time‑of‑day‑specific challenges.
    \item To the best of our knowledge, this is the first EPF study to evaluate forecasting models across all five regions of the NEM, enabling a rigorous assessment of model generalisability.
\end{enumerate}
The remainder of this paper is organised as follows. Section~\ref{sec2} reviews the relevant literature on DL-based EPF and advancements in time-series forecasting. Section~\ref{sec3} describes the methodology. Section~\ref{sec4} presents and analyses the results. Section~\ref{sec5} concludes the paper and outlines directions for future research.

\section{Related Work}\label{sec2}

\begin{table*}[!t]
\centering
\caption{Recently proposed time-series forecasting models and their key innovations.}
\label{tab:recent_models}
\begin{tabular}{llll}
\hline
\textbf{Model} & \textbf{Year} & \textbf{Family} & \textbf{Key Idea} \\
\hline
N-BEATS~\cite{Oreshkin2020-pe} & 2020 & MLP & Residual MLP blocks; learnable basis for trend and seasonality. \\
Informer~\cite{Zhou2021-yd} & 2021 & Transformer  & ProbSparse self-attention; attention distillation for long sequences. \\
Autoformer~\cite{Wu2021-sp} & 2021 & Transformer & Auto-correlation attention; built-in trend-seasonal decomposition. \\
SCINet~\cite{Liu2022-ar} & 2022 & CNN  & Recursive downsample–convolve–interact architecture. \\
S4~\cite{Gu2022-yb} & 2022 & SSM & Structured state-space layers for long-range dependencies. \\
FEDformer~\cite{Zhou2022-vq} & 2022 & Transformer & Fourier transform-based attention; efficient long-range modelling. \\
PatchTST~\cite{Nie2023-lh} & 2023 & Transformer & Patch tokenization; channel-independent attention mechanism. \\
DLinear~\cite{Zeng2023-zt} & 2023 & Linear  & Linear trend-seasonal decomposition with moving averages. \\
TimesNet~\cite{Wu2023-cm} & 2023 & CNN  & Converts 1D series to 2D temporal tensors to capture multi-period patterns. \\
iTransformer~\cite{Liu2024-sl} & 2023 & Transformer  & Variate-centric tokens; attention operates across variables. \\
Mamba~\cite{Gu2024-ve} & 2024 & SSM & Selective state-space layers enabling efficient long-sequence modeling. \\
TimeMixer~\cite{Wang2024-eg} & 2024 & MLP & Past-Decomposable Mixing and Future-Multipredictor-Mixing.\\
TimeXer~\cite{Wang2024-iy} & 2024 & Transformer & Patch-wise self-attention and variate-wise cross-attention. \\
\hline
\end{tabular}
\end{table*}

\subsection{Electricity Price Forecasting with Deep Learning}\label{sec2_1}
EPF has been extensively studied across various forecasting approaches, with recent developments increasingly centered on DL.
Below, we review DL-based EPF studies. In particular, the literature is analysed with respect to the temporal granularity of price data, the forecasting horizons considered, and the evolution of modelling architectures.

Whilst existing EPF studies exhibit variation in data granularity,  hourly data remain the predominant choice across diverse markets~\cite{Keles2016-tr, Li2021-bq, Prakash2023-to, Kuo2018-dk, Huang2021-zb, Heidarpanah2023-ax, Li2024-yh}. 
To better capture intraday volatility, some studies adopt finer temporal resolutions, including half-hourly~\cite{Tan2023-ck, Ghimire2024-ou, Li2025-zw}, 15-minute~\cite{Wang2025-mt}, and even 5-minute data~\cite{Deng2022-qo}. Despite differences in data resolution, the forecasting horizon in the literature 
primarily focused on hour-ahead~\cite{Khan2025-mj, Perla2025-he}, 2–3 hours ahead, and day-ahead settings~\cite{Lago2018-sh, Li2022-nq, Mubarak2024-rs}. 

Methodologically, recent EPF research reflects diverse strategies for modelling nonlinear and time-dependent patterns in electricity prices. Early studies applied feedforward neural architectures such as Artificial Neural Networks~\cite{Keles2016-tr} and Deep Neural Networks~\cite{Lago2018-sh}, establishing the viability of data-driven approaches for day-ahead forecasting. Subsequent work increasingly adopted sequence-based models, particularly Long Short-Term Memory (LSTM) networks~\cite{Li2021-bq}, often enhanced through signal decomposition techniques. Examples include wavelet–LSTM models~\cite{Chang2019-gg} and adaptive wavelet switching to accommodate evolving temporal characteristics~\cite{Iwabuchi2022-ou}. More advanced decomposition hybrids have also been proposed, such as Maximal Overlap Discrete Wavelet Transform–Empirical Mode Decomposition–LSTM~\cite{Prakash2023-to}. Parallel lines of work incorporate Gated Recurrent Unit (GRU)-based architectures and hybrid pipelines~\cite{Huang2021-zb, Li2024-yh}. Convolutional–recurrent combinations remain prominent~\cite{Kuo2018-dk, Heidarpanah2023-ax, Mubarak2024-rs}, while recent studies increasingly explore attention mechanisms and spatial learning, including Convolutional Neural Network (CNN)–self-attention~\cite{Pourdaryaei2024-kb}, CNN–GRU-attention~\cite{Ehsani2024-cz}, and attention-enhanced Graph Convolutional Networks~\cite{Yang2022-wu}. Transformer-based models have similarly emerged, such as Transformer–Bidirectional LSTM~\cite{Khan2025-mj} and hybrid GCN–Transformer frameworks~\cite{Wang2025-mt}, underscoring a growing shift toward attention-driven architectures in EPF.

\subsection{Recent Advancements in Time Series Forecasting}\label{sec2_2}
Research in time series forecasting has increasingly focused on improving the modelling of complex temporal patterns, including long-range dependencies, periodicity, and multivariate interactions. As shown in Table~\ref{tab:recent_models}, recent developments span several architectural families, including linear models, multilayer perceptron (MLP)-based, Transformer-based, CNN-based, and state-space models (SSMs), each introducing distinct mechanisms designed to enhance forecasting performance.

\paragraph{Linear and MLP-based models}
MLP‑based models such as N‑BEATS~\cite{Oreshkin2020-pe} demonstrated that simple feed‑forward architectures can remain competitive through explicit trend–seasonality decomposition. In parallel, the linear model DLinear~\cite{Zeng2023-zt} extended this idea using a single-layer linear architecture that decomposes inputs into trend and seasonal components before applying linear projections. More recent MLP variants, including TimeMixer~\cite{Wang2024-eg}, further operationalise multi-scale temporal modelling through specialised mixing modules. Together, these models highlight how both linear and lightweight MLP architectures can deliver strong forecasting performance while maintaining computational efficiency.

\paragraph{Transformer-based Models}
While self-attention provides strong global dependency modelling, its quadratic complexity has inspired a range of efficiency-focused designs. Informer~\cite{Zhou2021-yd}, Autoformer~\cite{Wu2021-sp}, and FEDformer~\cite{Zhou2022-vq} introduce sparse attention, decomposition-based attention, and frequency-domain operations, respectively. More recent variants such as PatchTST~\cite{Nie2023-lh}, iTransformer~\cite{Liu2024-sl}, and TimeXer~\cite{Wang2024-iy} improve the modelling of local structure, variable interactions, and exogenous inputs through patching strategies and variate-centric attention.

\paragraph{SSM and CNN-based Models}
SSM-based models offer an alternative to attention mechanisms by modelling continuous-time dynamics with linear complexity. S4~\cite{Gu2022-yb} established the foundation for long-range modelling through structured state-space layers, while Mamba~\cite{Gu2024-ve} introduced selective state updates that achieve attention-like selectivity with significantly lower computational overhead. CNN‑based models such as SCINet~\cite{Liu2022-ar} and TimesNet~\cite{Wu2023-cm} extract temporal patterns using recursive or hierarchical convolutions, with SCINet capturing multi‑resolution features and TimesNet modeling multi‑period structures through 2D temporal transformations.

\section{Methodology}\label{sec3}

\begin{figure*}[t!]
\centering
\includegraphics[width=\linewidth]{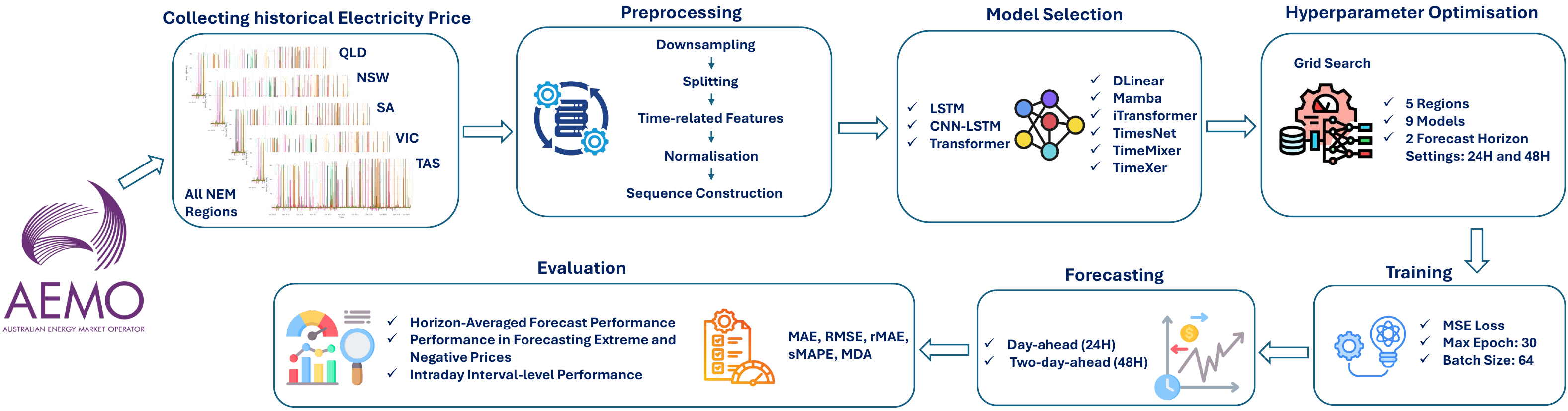}
\vspace{-\baselineskip}
\caption{An overview of the proposed EPF evaluation framework, encompassing stages from data collection to performance evaluation.}\label{fig1}
\end{figure*}

We propose an EPF evaluation framework, illustrated in Figure~\ref{fig1}, designed to systematically evaluate SOTA deep time-series models. 
The source code, datasets, and supplementary materials associated with the experiments in this article are publicly available on GitHub\footnote{\url{https://github.com/GaniMosman/Multi-Horizon-EPF-NEM.git}}.
The full workflow from data acquisition to model evaluation is described in the following subsections.

\subsection{Datasets}
For a comprehensive evaluation, we use data from all five regions of the NEM in Australia: Queensland (QLD), New South Wales (NSW) (including the Australian Capital Territory), Victoria (VIC), South Australia (SA), and Tasmania (TAS). The data correspond to the Regional Reference Price (RRP), which is the market-clearing price in each region. We retrieved the data from the Australian Energy Market Operator (AEMO)\footnote{\url{https://www.aemo.com.au/}} using the \texttt{nemdatatools} Python library\footnote{\url{https://pypi.org/project/nemdatatools/}}. Figure~\ref{tab:summary_stats} presents the summary statistics of the collected data. The dataset covers the period from January 1, 2023, to June 30, 2025, with a 5-minute temporal resolution (262,656 observations per region). Negative and extreme price observations are retained to preserve genuine market dynamics, and no smoothing or filtering is applied.

A key reason for selecting this period is that it reflects the NEM's most recent operational and structural conditions. This time window captures the post-COVID market environment, where demand patterns have stabilised following pandemic-related disruptions. It also coincides with a rapid increase in renewable penetration, which has fundamentally reshaped intraday price formation and contributed to negative prices, steeper ramps, and higher short-term volatility\footnote{\url{https://www.energycouncil.com.au/analysis/renewables-in-the-nem-are-they-leading-to-price-extremes}}. Finally, the 5-minute settlement rule, fully implemented in late 2021, was well established during this period, ensuring that the data represent the current dispatch and pricing structure. Using recent data, therefore, enables the models to be evaluated under contemporary market dynamics rather than historical conditions that may no longer be representative. 
\subsection{Data Preparation}
\paragraph{Downsampling}
AEMO operates a 5‑minute settlement system, with dispatch and spot prices set at 5‑minute intervals. For this study, prices were aggregated to a 30‑minute resolution to match the temporal scale of medium‑term operational analysis and to maintain computational tractability for multi‑step forecasting. Each 30‑minute value was obtained by averaging six consecutive 5‑minute RRPs, applied consistently across all five NEM regions. This reduced each regional dataset from 262,656 to 43,776 observations. As the original series contained no missing timestamps or values, no imputation was required.

\paragraph{Data Splitting}
The datasets are chronologically divided into training, validation, and testing subsets. Data from January 1, 2023, to December 31, 2024, are used for model training and validation, with 70\% allocated to training and the remaining 30\% for validation. The most recent six-month period, from January 1, 2025, to June 30, 2025, is held out for testing and inference. This forward-looking temporal split ensures that future information is not leaked into the training process, thereby providing a realistic evaluation of the model’s forecasting capability. 

\paragraph{Time-related Features}
The SOTA time series forecasting models evaluated in this study require time-related features as part of their input representations. Accordingly, four temporal features were extracted from the timestamps. These include hour of day, day of week, day of month, and month of year. These features are incorporated only into the SOTA time series models, as they are integral to the design of these architectures.

\paragraph{Normalisation}
Prior to model training, all input features are normalised using the Min–Max scaling method to the range $[0,1]$. Normalisation facilitates faster convergence in gradient-based models and helps maintain numerical stability during training. This approach is widely adopted in recent EPF studies.
\paragraph{Sequence Construction}
The normalised price series are transformed into input–output sequences for two experimental settings:
\begin{enumerate}
    \item A 7-day lookback window (336 half-hourly observations) used to forecast the next 24 hours (48 steps).
    \item A 14-day lookback window (672 half-hourly observations) used to forecast the next 48 hours (96 steps).
\end{enumerate}
All models adopt a direct multi‑horizon forecasting strategy. In each forecasting setting, a single forward pass outputs all 48 steps for the day‑ahead horizon and 96 steps for the two‑day‑ahead horizon.

\begin{table*}[t!]
\centering
\caption{Statistical summary of RRP at 5-min and 30-min intervals. Note: $\mu$: mean; $\sigma$: standard deviation; Skew: skewness; Kurt: kurtosis.}
\label{tab:summary_stats}
\resizebox{\textwidth}{!}{
\begin{tabular}{@{} l | *{14}{r} @{}}
\toprule
\textbf{Dataset} & \multicolumn{7}{c}{\textbf{5-min Interval}} & \multicolumn{7}{c}{\textbf{30-min Interval}} \\
\cmidrule(lr){2-8} \cmidrule(lr){9-15}
 & $\mu$ & $\sigma$ & Min & 50\% & Max & Skew & Kurt &
   $\mu$ & $\sigma$ & Min & 50\% & Max & Skew & Kurt \\
\midrule
QLD & 101.947 & 420.073 & -801.064 & 83.267 & 17500.000 & 30.228 & 996.902 & 101.947 & 336.859 & -195.681 & 82.225 & 15631.376 & 27.895 & 959.286 \\
NSW & 115.674 & 545.631 & -999.998 & 83.890 & 17500.000 & 25.424 & 688.095 & 115.674 & 459.463 & -683.954 & 82.752 & 17500.000 & 24.579 & 697.665 \\
VIC & 74.544 & 332.302 & -1000.000 & 61.793 & 17500.000 & 37.872 & 1644.046 & 74.544 & 310.363 & -686.552 & 62.298 & 16600.000 & 36.565 & 1590.184 \\
SA & 92.758 & 484.672 & -999.999 & 68.328 & 17500.000 & 25.057 & 707.655 & 92.758 & 413.774 & -945.548 & 68.985 & 16969.981 & 23.398 & 671.295 \\
TAS & 85.777 & 251.013 & -998.068 & 70.200 & 17500.000 & 48.087 & 2709.027 & 85.777 & 216.386 & -922.399 & 71.394 & 16600.000 & 46.256 & 2751.942 \\
\bottomrule
\end{tabular}
}
\end{table*}

\subsection{Model Selection}
In this section, we explain how we selected the deep time series models included in our evaluation, along with the models commonly used in existing EPF research, which serve as standard DL baselines.  These baseline models enable us to compare the performance of the deep time series models with those already applied in the EPF literature. Below, we outline our model selection strategy and the rationale behind it.

\paragraph{Models from Existing EPF Research} 
The comprehensive review of existing EPF studies in Section~\ref{sec2_1} revealed that among recurrent models, LSTM and GRU are the widely applied, with LSTM being the most frequently used. Accordingly, we selected LSTM as the representative recurrent model for this study.
Hybrid DL models that combine convolutional and recurrent structures, particularly CNN-LSTM, are also widely applied in the EPF literature; therefore, we included CNN–LSTM as a representative hybrid architecture.
In addition, recent EPF studies have increasingly utilised transformer-based architectures, motivated by their ability to model long-term dependencies via self-attention mechanisms. For this reason, the Transformer model was also incorporated into our evaluation. This study deliberately excludes signal decomposition to ensure that performance reflects the inherent capability of each DL architecture rather than the influence of preprocessing strategies. Together, LSTM, CNN-LSTM, and Transformer represent the dominant DL model families explored in existing EPF research covering recurrent, hybrid, and attention-based models. However, a noted challenge in the EPF field is the limited reproducibility of existing publications. Given that exact replication of prior implementations was not feasible, we constructed our own representative implementations of these architectures.

\paragraph{Models from Recent Time Series Forecasting Research}
We reviewed SOTA models from the time series forecasting literature (see Section~\ref{sec2_2}). Numerous models have been proposed in recent years; a subset of the most influential ones is summarised in Table~\ref{tab:recent_models}, categorised based on their underlying algorithmic family.
Following the benchmarking leaderboard\footnote{\url{https://github.com/thuml/Time-Series-Library}} that ranks recent long-term forecasting models, we selected two top-performing models, TimeXer and TimeMixer, representing the lookback-searching and lookback-96 subcategories, respectively.
Furthermore, to ensure coverage across major model families, we selected one representative model from each: TimesNet (CNN family), iTransformer (transformer family), and Mamba (SSM family). In addition to deep models, we also include DLinear, a lightweight linear model. The original DLinear study reported competitive and, in many cases, superior performance to transformer‑based models on standard long‑term forecasting benchmarks, motivating its inclusion in this study. The detailed mathematical formulations of these models are skipped in this paper for brevity. We refer the reader to the original articles (DLinear~\cite{Zeng2023-zt}, TimesNet~\cite{Wu2023-cm}, iTransformer~\cite{Liu2024-sl}, Mamba~\cite{Gu2024-ve}, TimeMixer~\cite{Wang2024-eg}, and TimeXer~\cite{Wang2024-iy}) for their technical descriptions.

\subsection{Hyperparameter Optimisation}
We conducted a comprehensive hyperparameter search across all models using grid search to ensure fair performance comparisons.
The hyperparameter search was conducted separately for each dataset and each forecasting horizon setting across five regions, resulting in $2$ forecasting horizons $\times$ $5$ datasets $= 10$ unique searches for each of the $9$ models. For the LSTM, Mamba, Transformer, iTransformer, TimesNet, TimeMixer, and TimeXer models, we optimised the model dimension (i.e., hidden units), the number of layers, and the learning rate (LR), yielding 50 combinations per search. DLinear has only one tunable hyperparameter, the LR, whereas for CNN-LSTM, we optimised the number of CNN filters, filter size, LSTM hidden dimension, and LR, resulting in 375 combinations per search. The hyperparameter search space for all models is presented in Table~\ref{tab:hyperparam_search_space}. The resulting optimal hyperparameters are provided on GitHub.

\begin{table*}[tb!]
\centering
\caption{Hyperparameter search space for all evaluated models.}
\label{tab:hyperparam_search_space}

\begin{tabular}{p{5cm}ccccc}  
\toprule
\textbf{Model(s)} & \textbf{Learning Rate} & \textbf{Dimensions} & \textbf{Layers} & \textbf{Filter Size} & \textbf{Filters} \\
\midrule
LSTM, Mamba, Transformer, iTransformer, TimesNet, TimeMixer, TimeExer
& 0.001, 0.005, 0.01, 0.05, 0.1 & 32, 64, 128, 256, 512 & 1--2 & N/A & N/A \\
\midrule
DLinear & 0.001, 0.005, 0.01, 0.05, 0.1 & N/A & N/A & N/A & N/A \\
\midrule
CNN-LSTM & 0.001, 0.005, 0.01, 0.05, 0.1 & 32, 64, 128, 256, 512 & N/A & 3, 5, 7 & 32, 64, 128, 256, 512 \\
\bottomrule
\end{tabular}
\end{table*}

\subsection{Experimental Setup}
We conducted all experiments in Python, primarily utilising the PyTorch framework. For the SOTA time series models, we collected the original source code from the authors' official GitHub repositories. All code was executed on NVIDIA H100 Tensor Core GPUs, within a resource allocation limit of 256 GB of system memory and 12 CPU cores. Training for all models was optimised using the Adam optimiser, with the Mean Squared Error as the loss function. We employed an LR scheduler, ReduceLROnPlateau, to dynamically adjust the rate during training. Models were trained for a maximum of 30 epochs with a batch size of 64, and we utilised an early stopping mechanism with a patience of 10 epochs based on the validation loss. For the reported results, we conducted each experiment five (5) times using different random seeds. 
\subsection{Evaluation}
This study evaluates model performance across three dimensions.~\emph{First}, we assess overall performance by computing forecast errors averaged across all look‑ahead steps, providing a holistic measure of each model’s predictive capability.~\emph{Second}, we evaluate model behaviour under extreme market conditions by examining forecast accuracy within the upper and lower 5\% of the price distribution, offering insight into robustness during rare but impactful events.~\emph{Third}, We analyse intraday performance by assessing forecast accuracy at each of the 48 half‑hour intervals, providing insight into time‑of‑day‑dependent variations in model behaviour.

\subsubsection{Metrics}
We evaluate the models' performance using Mean Absolute Error (MAE), Root Mean Square Error (RMSE), Relative Mean Absolute Error (rMAE), Symmetric Mean Absolute Percentage Error (sMAPE), and Mean Directional Accuracy (MDA). Standard definitions of MAE, RMSE, and sMAPE are omitted for brevity.
rMAE is defined as:
\begin{equation}
\text{rMAE} = \frac{\text{MAE}_{\text{model}}}{\text{MAE}_{\text{benchmark}}} =
\frac{ \frac{1}{N} \sum_{i=1}^{N} \left| y_i - \hat{y}_i \right| }
{ \frac{1}{M-f} \sum_{t=f+1}^{M} \left| y_t - y_{t-f} \right| }
\end{equation}
where $N$ is the total number of forecast points, $y_i$ is actual observed price, and $\hat{y}_i$ is predicted price. The denominator is the MAE of a weekly seasonal na\"{i}ve forecast ($f = 336$ for 30-minute data); $y_t$ denotes the actual observed price at time $t$, and $y_{t-f}$ represents the observed price one week earlier, as recommended by~\cite{Lago2021-oj}. Here, $f$ denotes the seasonal frequency of the data, and $M$ denotes the total number of out-of-sample observations used to compute the benchmark error. 

MDA for multi-step forecasts is defined as:
\begin{equation}
\text{MDA} = \frac{1}{N-1} \sum_{t=2}^{N} \mathbf{1} \Big[ \text{sign}(\hat{y}_t - \hat{y}_{t-1}) = \text{sign}(y_t - y_{t-1}) \Big] \times 100
\end{equation}
where \(\mathbf{1}[\cdot]\) is an indicator function equal to 1 if the predicted change direction matches the actual change direction, and 0 otherwise.



\section{Results and Discussion}\label{sec4}
\subsection{Overall Model Performance}\label{sec:performance}

\begin{table*}[!t]
\caption{Models performance on all five regions across 24-hour and 48-hour forecast horizons. Lower values indicate better performance, except for MDA, where higher values are preferred. MAE and RMSE are reported in Australian dollars (A\$/MWh). Best and second-best results are indicated in \textbf{bold} and \underline{underline}, respectively.}
\label{tab:all_metrics}
\centering
\begin{tabular}{@{} c  c | *5r | *5r @{}}
\toprule
\textbf{Region} & \textbf{Model} & \multicolumn{5}{c}{\textbf{24H}} & \multicolumn{5}{c}{\textbf{48H}} \\
\cmidrule(lr){3-12}
& & MAE & RMSE & sMAPE & rMAE & MDA & MAE & RMSE & sMAPE & rMAE & MDA \\ \midrule
QLD & CNN-LSTM & 59.031 & 393.466 & 57.575 & 0.744 & \textbf{66.184} & \textbf{64.178} & \textbf{407.157} & \textbf{62.298} & \textbf{0.808} & \textbf{65.574} \\
& DLinear & 77.637 & 408.145 & 68.122 & 0.978 & 54.644 & 91.225 & 424.292 & 78.473 & 1.149 & 54.039 \\
& LSTM & \underline{57.875} & 393.397 & \textbf{56.145} & \underline{0.729} & \underline{65.358} & \underline{64.477} & \underline{408.129} & \underline{65.535} & \underline{0.812} & 64.684 \\
& Mamba & 58.696 & \underline{392.834} & \underline{57.027} & 0.739 & 62.951 & 71.461 & 425.951 & 65.652 & 0.9 & 63.5 \\
& TimeMixer & 88.805 & 498.688 & 64.259 & 1.119 & 59.862 & 89.398 & 429.763 & 71.812 & 1.126 & 57.093 \\
& TimeXer & 65.849 & 396.833 & 61.959 & 0.829 & 59.778 & 76.23 & 418.272 & 68.122 & 0.96 & 59.865 \\
& TimesNet & 75.25 & 490.291 & 63.128 & 0.948 & 63.298 & 83.498 & 472.201 & 70.736 & 1.052 & 63.184 \\
& Transformer & \textbf{54.938} & \textbf{386.188} & 57.542 & \textbf{0.692} & 64.536 & 64.802 & 410.117 & 65.7 & 0.816 & \underline{65.010} \\
& iTransformer & 75.475 & 443.239 & 63.822 & 0.951 & 59.033 & 76.922 & 422.678 & 69.505 & 0.969 & 60.505 \\
\midrule
NSW & CNN-LSTM & \textbf{82.409} & 513.079 & \textbf{56.104} & \textbf{0.798} & \textbf{61.397} & 96.16 & 584.364 & 61.991 & 0.931 & \textbf{62.851} \\
& DLinear & 102.954 & 515.743 & 66.8 & 0.997 & 53.817 & 98.861 & \underline{518.314} & 67.805 & 0.958 & 53.97 \\
& LSTM & 86.414 & \textbf{511.609} & \underline{61.144} & 0.837 & 59.831 & 89.618 & 524.464 & \underline{61.814} & 0.868 & \underline{61.924} \\
& Mamba & \underline{82.448} & 518.251 & 61.385 & \underline{0.799} & 58.667 & \textbf{85.856} & 519.28 & 62.827 & \textbf{0.832} & 61.315 \\
& TimeMixer & 98.054 & 514.872 & 71.03 & 0.95 & 52.922 & 93.128 & 518.969 & 70.144 & 0.902 & 52.161 \\
& TimeXer & 89.044 & 519.295 & 63.446 & 0.862 & 57.533 & 93.506 & 519.353 & 73.65 & 0.906 & 54.895 \\
& TimesNet & 91.024 & 523.729 & 63.26 & 0.882 & 57.243 & 91.464 & 518.829 & 69.97 & 0.886 & 55.384 \\
& Transformer & 84.095 & \underline{511.923} & 62.983 & 0.815 & \underline{60.939} & \underline{89.388} & \textbf{517.518} & \textbf{61.015} & \underline{0.866} & 60.921 \\
& iTransformer & 89.408 & 517.652 & 66.079 & 0.866 & 57.073 & 89.951 & 520.873 & 66.305 & 0.871 & 57.336 \\
\midrule
SA & CNN-LSTM & \textbf{87.158} & 490.976 & \textbf{86.692} & \textbf{0.708} & \textbf{62.358} & 100.908 & 509.015 & 96.704 & 0.819 & 57.832 \\
& DLinear & 95.607 & \underline{487.599} & 89.818 & 0.776 & 53.947 & 99.231 & \textbf{499.648} & 93.248 & 0.806 & 54.741 \\
& LSTM & \underline{88.321} & 491.239 & 88.504 & \underline{0.717} & \underline{59.626} & 99.872 & 509.8 & 96.72 & 0.811 & 56.831 \\
& Mamba & 89.231 & 494.933 & 88.258 & 0.725 & 59.555 & \underline{96.045} & 505.923 & \underline{92.239} & \underline{0.780} & \textbf{59.743} \\
& TimeMixer & 96.21 & 490.463 & 88.494 & 0.781 & 57.432 & 103.727 & 504.528 & 94.523 & 0.842 & 54.135 \\
& TimeXer & 90.984 & \textbf{484.042} & \underline{87.400} & 0.739 & 59.055 & \textbf{93.963} & \underline{499.855} & \textbf{90.905} & \textbf{0.763} & 56.253 \\
& TimesNet & 96.504 & 489.439 & 93.473 & 0.784 & 57.928 & 98.355 & 501.358 & 95.296 & 0.799 & \underline{58.782} \\
& Transformer & 90.798 & 492.023 & 87.686 & 0.737 & 59.328 & 100.309 & 507.358 & 94.51 & 0.815 & 57.601 \\
& iTransformer & 95.388 & 487.948 & 91.298 & 0.775 & 57.122 & 101.108 & 502.003 & 95.966 & 0.821 & 58.371 \\
\midrule
TAS & CNN-LSTM & 46.562 & 296.75 & \underline{36.543} & 0.7 & 54.578 & 51.614 & 306.289 & 40.205 & 0.776 & 53.78 \\
& DLinear & 53.765 & \textbf{296.015} & 42.538 & 0.809 & 51.334 & 57.997 & 306.038 & 46.357 & 0.872 & 51.179 \\
& LSTM & \underline{46.443} & 297.064 & 37.046 & \underline{0.699} & \textbf{56.297} & \underline{51.324} & \underline{305.927} & \underline{40.066} & \underline{0.772} & \textbf{54.775} \\
& Mamba & \textbf{46.071} & 297.836 & 36.565 & \textbf{0.693} & 54.546 & 53.569 & 308.344 & 41.715 & 0.806 & \underline{54.737} \\
& TimeMixer & 50.24 & 297.451 & 38.292 & 0.756 & 55.251 & 57.334 & 309.074 & 42.89 & 0.862 & 52.144 \\
& TimeXer & 46.783 & \underline{296.559} & \textbf{36.246} & 0.704 & 54.503 & \textbf{49.459} & \textbf{305.054} & \textbf{39.140} & \textbf{0.744} & 53.362 \\
& TimesNet & 48.768 & 299.065 & 37.449 & 0.733 & 55.283 & 52.82 & 307.377 & 41.715 & 0.794 & 53.646 \\
& Transformer & 48.263 & 299.326 & 37.361 & 0.726 & 53.675 & 51.753 & 306.654 & 40.234 & 0.778 & 52.913 \\
& iTransformer & 51.438 & 298.873 & 38.549 & 0.774 & \underline{55.288} & 53.886 & 307.291 & 42.175 & 0.81 & 50.0 \\
\midrule
VIC & CNN-LSTM & 75.013 & 467.766 & 83.663 & 0.74 & 61.965 & 86.816 & 476.73 & 92.741 & 0.856 & 54.954 \\
& DLinear & 80.312 & \textbf{455.523} & 85.856 & 0.792 & 56.553 & \underline{82.740} & \textbf{463.210} & 90.417 & \underline{0.816} & 58.038 \\
& LSTM & 75.213 & 471.492 & 83.876 & 0.742 & \underline{62.781} & 84.0 & 476.018 & 90.79 & 0.829 & \underline{61.070} \\
& Mamba & \underline{73.153} & 463.357 & 83.281 & \underline{0.722} & 62.58 & 89.305 & 481.541 & 91.519 & 0.881 & 59.446 \\
& TimeMixer & 80.238 & 459.176 & 80.919 & 0.791 & 60.843 & 89.804 & 477.208 & \underline{87.196} & 0.886 & 59.833 \\
& TimeXer & \textbf{71.859} & \underline{458.037} & \textbf{77.753} & \textbf{0.709} & 60.375 & \textbf{78.752} & 468.272 & \textbf{84.690} & \textbf{0.777} & 59.161 \\
& TimesNet & 77.484 & 462.307 & \underline{80.475} & 0.764 & \textbf{63.338} & 85.632 & \underline{464.397} & 89.019 & 0.845 & 59.926 \\
& Transformer & 73.454 & 463.728 & 85.939 & 0.725 & 60.713 & 83.124 & 476.059 & 90.501 & 0.82 & 59.299 \\
& iTransformer & 82.177 & 465.745 & 82.334 & 0.811 & 61.442 & 88.15 & 467.769 & 87.282 & 0.87 & \textbf{62.453} \\
\midrule
Average & CNN-LSTM & \underline{70.035} & 432.407 & \textbf{64.115} & \underline{0.738} & \textbf{61.297} & 79.935 & 456.711 & \underline{70.788} & 0.838 & 58.998 \\
& DLinear & 82.055 & 432.605 & 70.627 & 0.870 & 54.059 & 86.011 & \underline{442.300} & 75.260 & 0.920 & 54.393 \\
& LSTM & 70.853 & 432.960 & 65.343 & 0.745 & \underline{60.778} & \textbf{77.858} & 444.867 & 70.985 & \textbf{0.818} & \textbf{59.857} \\
& Mamba & \textbf{69.920} & 433.442 & \underline{65.303} & \textbf{0.735} & 59.660 & 79.247 & 448.208 & 70.790 & 0.840 & \underline{59.748} \\
& TimeMixer & 82.710 & 452.130 & 68.599 & 0.879 & 57.262 & 86.678 & 447.908 & 73.313 & 0.924 & 55.073 \\
& TimeXer & 72.904 & \underline{430.953} & 65.361 & 0.769 & 58.249 & 78.382 & \textbf{442.161} & 71.301 & 0.830 & 56.707 \\
& TimesNet & 77.806 & 452.966 & 67.557 & 0.822 & 59.418 & 82.354 & 452.833 & 73.347 & 0.875 & 58.184 \\
& Transformer & 70.310 & \textbf{430.638} & 66.302 & 0.739 & 59.838 & \underline{77.875} & 443.541 & \textbf{70.392} & \underline{0.819} & 59.149 \\
& iTransformer & 78.777 & 442.691 & 68.416 & 0.835 & 57.992 & 82.003 & 444.123 & 72.247 & 0.868 & 57.733 \\
\bottomrule
\end{tabular}
\end{table*}

Table~\ref{tab:all_metrics} reports model performance across the five NEM regions for 24-hour and 48-hour forecast horizons. The results show that standard DL baselines (LSTM, CNN‑LSTM, Transformer) often outperform SOTA time‑series architectures across many regions. CNN-LSTM and LSTM frequently attain first or second place across MAE, RMSE, sMAPE, and MDA, including in volatile markets such as SA and NSW. Although prior work (e.g., DLinear~\cite{Zeng2023-zt}) reported that simple linear layers can outperform transformers, our results show the opposite: DLinear (linear) and TimeMixer (MLP-based) generally underperform, with rMAE occasionally exceeding 1, indicating worse performance than a weekly naive benchmark. That said, DLinear is competitive on RMSE in TAS and VIC, including the best RMSE in VIC at both forecast horizons. TimesNet and iTransformer rarely appear among the top performers. Certain newer architectures exhibit region-specific strengths: TimeXer is dominant in VIC—particularly at the 48-hour horizon, where it achieves first or second place across most metrics—and Mamba is competitive in QLD and SA.

The average results reinforce these observations. For the 24-hour horizon, CNN-LSTM and Transformer rank among the strongest-performing models across most metrics, while Mamba is the only SOTA model to achieve top performance in selected metrics. DLinear and TimeMixer rank among the weakest-performing models on average. At the 48-hour horizon, errors increase across all models; however, LSTM and Transformer remain top-ranked across most metrics, with TimeXer being the only SOTA model that consistently challenges these baselines.

Directional accuracy, as measured by MDA, further distinguishes the baseline models. CNN‑LSTM consistently achieves high MDA across regions and horizons, indicating a strong ability to capture the direction of price movements, while LSTM is often competitive. This is particularly important in contexts where correctly predicting the direction of price changes can be more valuable than minimising pointwise error alone. Error magnitudes also vary substantially across regions: SA and VIC exhibit the highest error levels, while TAS consistently records the lowest.

In summary, despite their sophisticated architectural design, the results demonstrate that SOTA time‑series models are often outperformed by simple DL baselines in the highly volatile context of electricity markets.

\begin{table}[t]
\caption{Performance of models on extreme-price observations, evaluated using the upper 5\% and lower 5\% tails of the distribution. Best and second-best results are indicated in \textbf{bold} and \underline{underline}, respectively.
}
\label{tab:extreme_price}
\centering
\begingroup
\scriptsize
\setlength{\tabcolsep}{3pt}
\renewcommand{\arraystretch}{0.95}
\resizebox{\columnwidth}{!}{%
\begin{tabular}{@{} l @{\hspace{6pt}} *3{r} | *3{r} @{}}
\toprule
\textbf{Model} & \multicolumn{3}{c}{\textbf{24H}} & \multicolumn{3}{c}{\textbf{48H}} \\
\cmidrule(lr){2-4} \cmidrule(lr){5-7}
 & MAE & RMSE & sMAPE & MAE & RMSE & sMAPE \\
\midrule
CNN-LSTM & 340.947 & 1380.162 & 110.733 & 360.599 & 1424.722 & 126.286 \\
DLinear & 371.167 & \underline{1370.482} & 127.684 & 369.037 & \textbf{1387.676} & 128.611 \\
LSTM & \underline{337.224} & 1381.626 & \underline{108.640} & 355.485 & 1403.804 & 125.705 \\
Mamba & 342.208 & 1382.109 & 111.024 & 359.738 & 1409.938 & 125.829 \\
TimeMixer & 359.319 & 1384.998 & 114.199 & 366.947 & 1397.790 & 125.883 \\
TimeXer & \textbf{330.247} & \textbf{1368.385} & \textbf{106.478} & \textbf{346.588} & 1394.403 & \textbf{115.216} \\
TimesNet & 355.009 & 1409.726 & 116.139 & 369.320 & 1409.654 & 129.950 \\
Transformer & 337.407 & 1376.974 & 118.136 & \underline{355.111} & 1404.712 & 128.728 \\
iTransformer & 356.088 & 1393.305 & 118.760 & 357.569 & \underline{1393.066} & \underline{121.822} \\
\bottomrule
\end{tabular}
}
\endgroup
\end{table}

\subsection{Performance on Extreme and Negative Prices}\label{sec:performance_EX_n_Neg}

Tables~\ref{tab:extreme_price} and~\ref{tab:neg_price} report model performance during extreme and negative price events. The results show that, on the upper and lower 5\% tails of the price distribution, absolute and relative errors increase substantially for all models, confirming that rare spikes remain challenging to forecast. TimeXer consistently performs best, achieving the lowest errors at the 24-hour horizon and maintaining the lowest MAE and sMAPE at 48 hours. LSTM frequently achieves second-best performance across most metrics at the 24-hour horizon, remaining highly competitive with SOTA time-series models. Mamba shows competitive performance, while DLinear occasionally achieves strong RMSE, suggesting some robustness to large absolute deviations despite weaker overall performance. In contrast, TimeMixer and TimesNet consistently rank among the weakest-performing models, while iTransformer shows competitive performance on the 48-hour forecast horizon.

Negative-price periods further expose model limitations. TimeXer again leads on MAE and sMAPE for both horizons, but sMAPE remains very high across all models, reflecting instability around zero and frequent sign errors. Baseline models such as LSTM and CNN-LSTM remain competitive on MAE and RMSE, while Transformer tends to underperform in this regime. Crucially, none of the other deep SOTA architectures outperform these standard baselines under negative-price conditions, indicating that neither traditional DL nor recent SOTA designs adequately address the unique dynamics of negative pricing.

\begin{table}[t]
\caption{Performance of models on negative-price observations. 
Best and second-best results are indicated in \textbf{bold} and \underline{underline}, respectively.
}
\label{tab:neg_price}
\centering
\begingroup
\scriptsize
\setlength{\tabcolsep}{3pt}
\renewcommand{\arraystretch}{0.95}
\resizebox{\columnwidth}{!}{%
\begin{tabular}{@{} l @{\hspace{6pt}} *3{r} | *3{r} @{}}
\toprule
\textbf{Model} & \multicolumn{3}{c}{\textbf{24H}} & \multicolumn{3}{c}{\textbf{48H}} \\
\cmidrule(lr){2-4} \cmidrule(lr){5-7}
 & MAE & RMSE & sMAPE & MAE & RMSE & sMAPE \\
\midrule
CNN-LSTM & \underline{69.305} & \underline{85.524} & 178.663 & 83.959 & 96.189 & 189.930 \\
DLinear & 89.235 & 104.340 & 191.543 & 86.843 & 103.551 & 187.900 \\
LSTM & 69.826 & \textbf{85.238} & 174.071 & \underline{81.886} & \textbf{93.106} & 187.208 \\
Mamba & 71.616 & 103.352 & 176.067 & 83.569 & 96.265 & 186.713 \\
TimeMixer & 74.608 & 105.114 & \underline{173.463} & 85.906 & 106.538 & 184.044 \\
TimeXer & \textbf{69.042} & 87.035 & \textbf{171.518} & \textbf{76.798} & \underline{94.228} & \textbf{177.242} \\
TimesNet & 75.573 & 90.926 & 180.664 & 85.147 & 97.402 & 189.457 \\
Transformer & 78.535 & 90.368 & 186.799 & 90.899 & 100.666 & 194.613 \\
iTransformer & 77.450 & 94.162 & 183.162 & 82.693 & 101.019 & \underline{182.415} \\
\bottomrule
\end{tabular}
}
\endgroup
\end{table}

\subsection{Intraday Interval-level Performance}
To gain deeper insights into the temporal dynamics of the price forecasting accuracy, we conducted a granular analysis of model performance across the 48 half-hour intervals of the day. By evaluating all the models at each time interval, we reveal that \textit{price forecasting difficulty is not uniform throughout the day} but clusters around specific periods driven by underlying market mechanics. All NEM regions show broadly comparable diurnal patterns in forecasting performance. Therefore, for illustrative clarity and analytical conciseness, we restrict our discussion to the QLD region as a representative case. Intraday performance visualisations for the remaining NEM regions are provided on GitHub.

\subsubsection{Absolute Errors Spike Driven by High Evening Volatility}\label{evening_volatility}
\begin{figure}[!t]
\centering


\subfloat[MAE of all models in the QLD region, evaluated at 30-minute intervals.]{%
  \includegraphics[width=\linewidth]{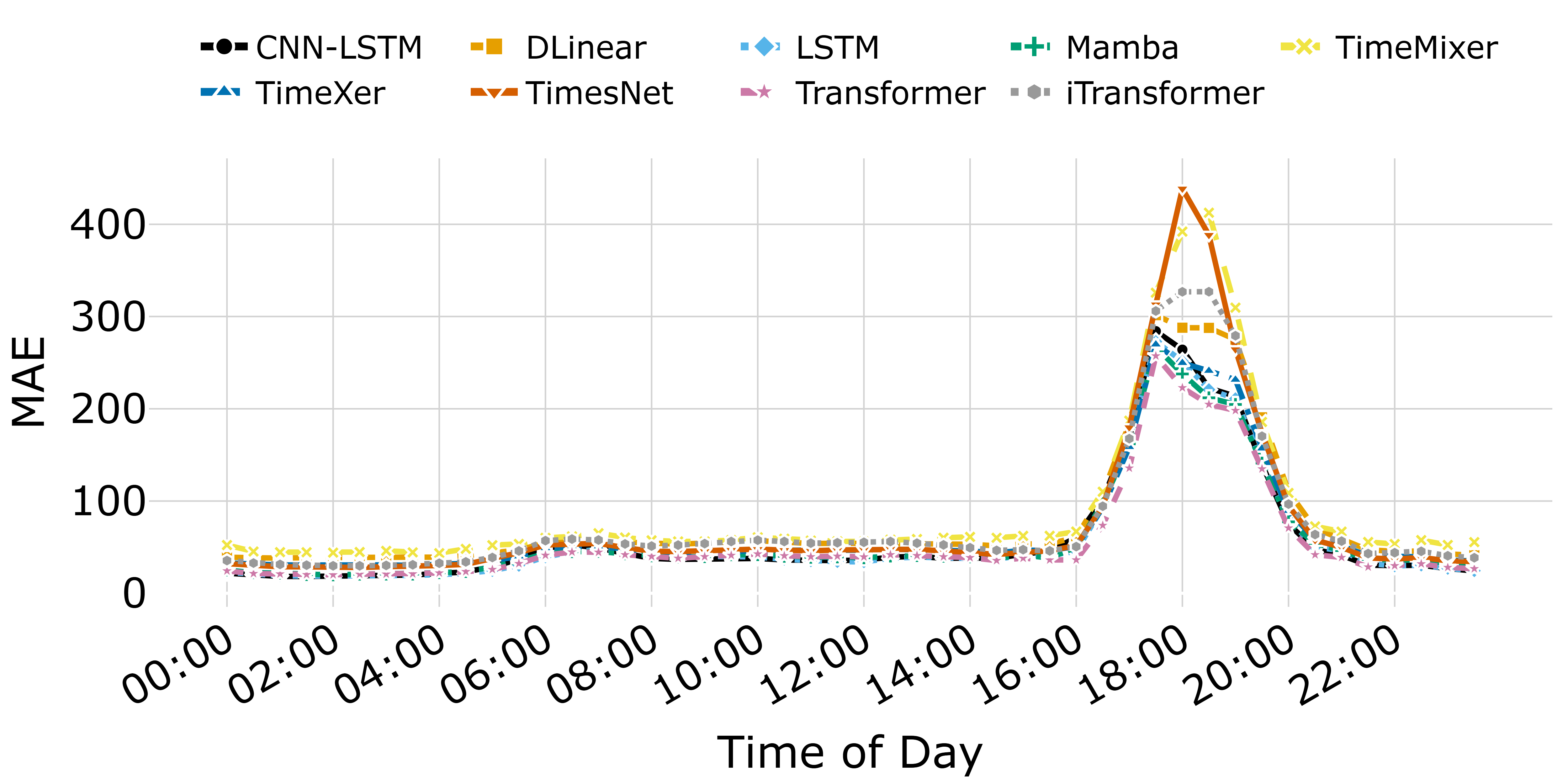}%
  \label{fig:qld_mae}%
}


\subfloat[RMSE of all models in the QLD region, evaluated at 30-minute intervals.]{%
  \includegraphics[width=\linewidth]{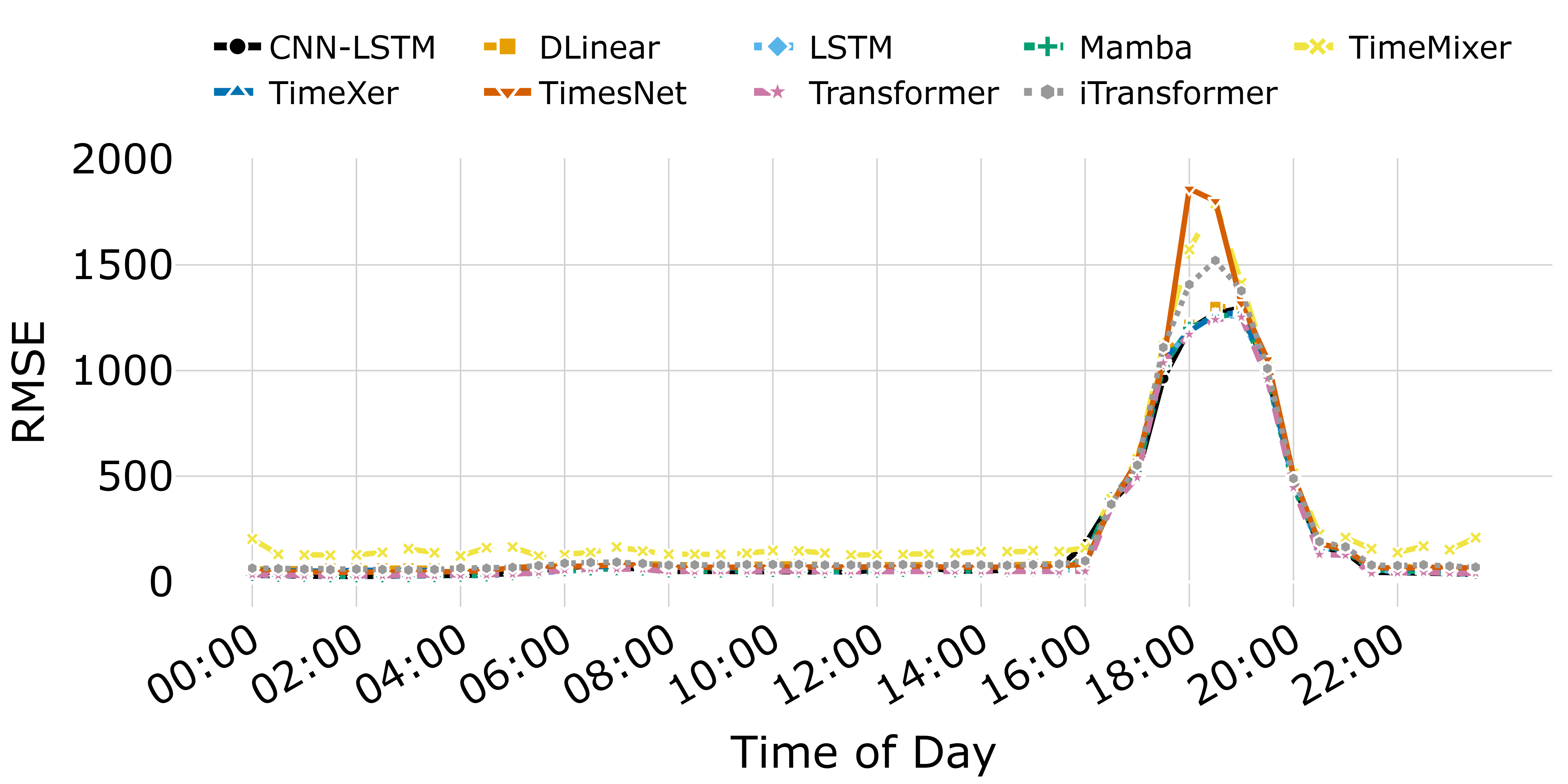}%
  \label{fig:qld_rmse}%
}

\subfloat[Diurnal electricity price dynamics in the QLD region, showing the standard deviation of price changes and the average price. The shaded region (16:00--20:30) highlights periods associated with elevated forecasting errors.]{%
  \includegraphics[width=\linewidth]{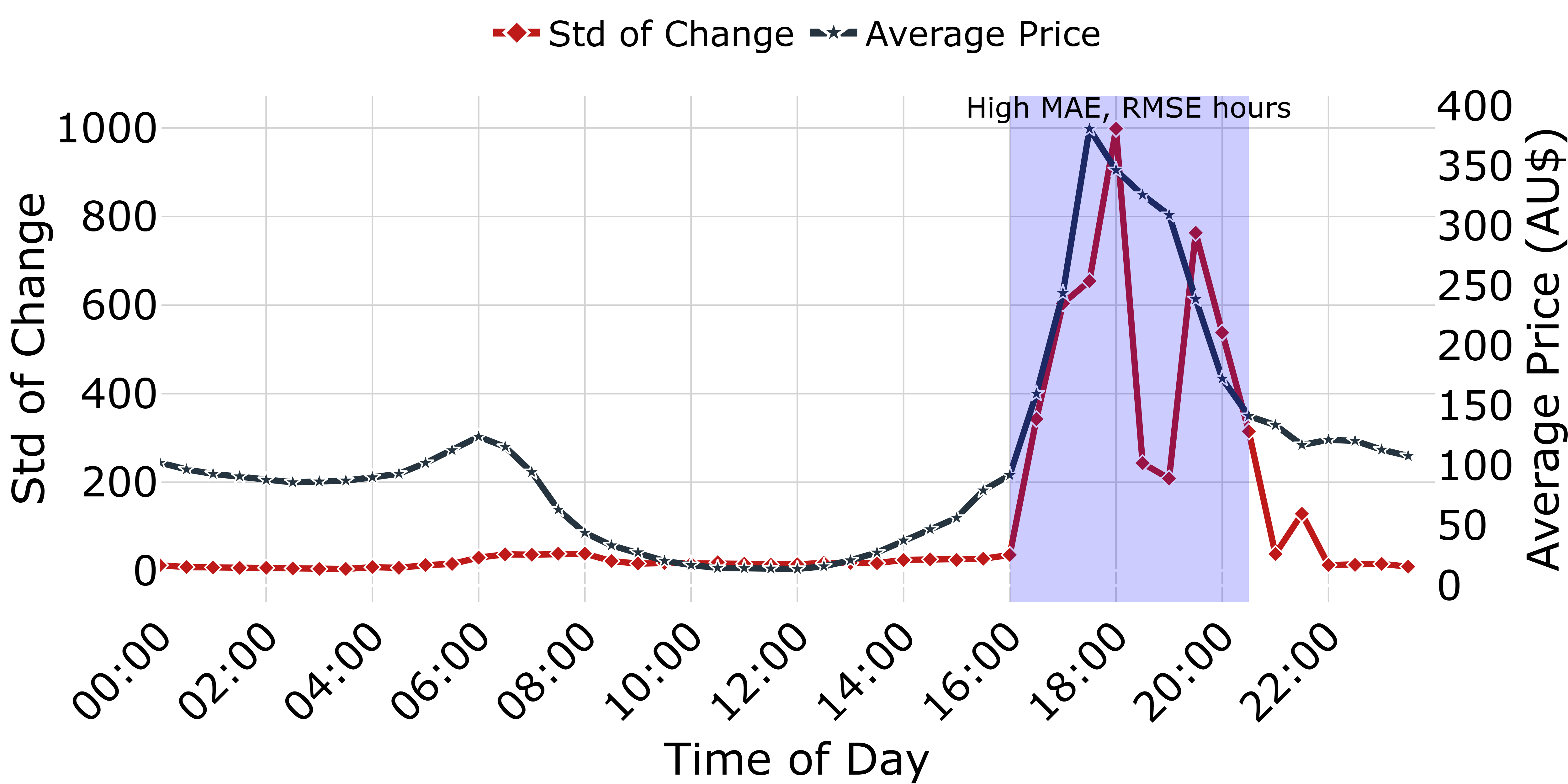}%
  \label{fig:qld_volatility}%
}
\caption{Intraday interval-level absolute forecasting errors and diurnal price characteristics for the \textbf{QLD} region.}
\label{fig:qld_three_panel}
\end{figure}

The absolute error metrics, MAE and RMSE (Figures~\ref {fig:qld_mae} and~\ref {fig:qld_rmse}), demonstrate performance degradation concentrated in specific high-volatility windows. As illustrated in Figure~\ref{fig:qld_volatility}, which plots the standard deviation of price changes (red line) alongside the average price (blue line) for each half-hour interval, error peaks in MAE and RMSE align closely with periods of elevated short-term price volatility. These high-volatility hours, spanning approximately 16:00 to 20:30, correspond to the evening ramp-up in demand as solar generation wanes, a hallmark of QLD’s solar-heavy profile. Although MAE and RMSE increase for all models during the evening volatility window, the rise is comparatively modest for sequence‑based architectures such as Transformers, Mamba, and LSTM, whereas models like TimeMixer, TimesNet, and others experience noticeably larger error escalations. During this transition, the standard deviation of price changes surges dramatically, reaching up to 1000~A\$/MWh.
This surge reflects abrupt shifts as dispatchable sources, including coal, gas, and battery discharges, ramp up to offset declining solar output.
This behaviour intensifies short-term volatility, making these hours inherently challenging and amplifying forecasting errors as models struggle to capture the resulting dynamics. The average price curve further contextualises this, showing a dip during solar peaks (around 7:00–15:00), reflecting the low marginal cost associated with abundant solar generation, followed by a rapid increase during the evening ramp-up.


\begin{figure}[!t]
\centering

\subfloat[sMAPE of all models in the QLD region, evaluated at 30-minute intervals. ]{%
  \includegraphics[width=\linewidth]{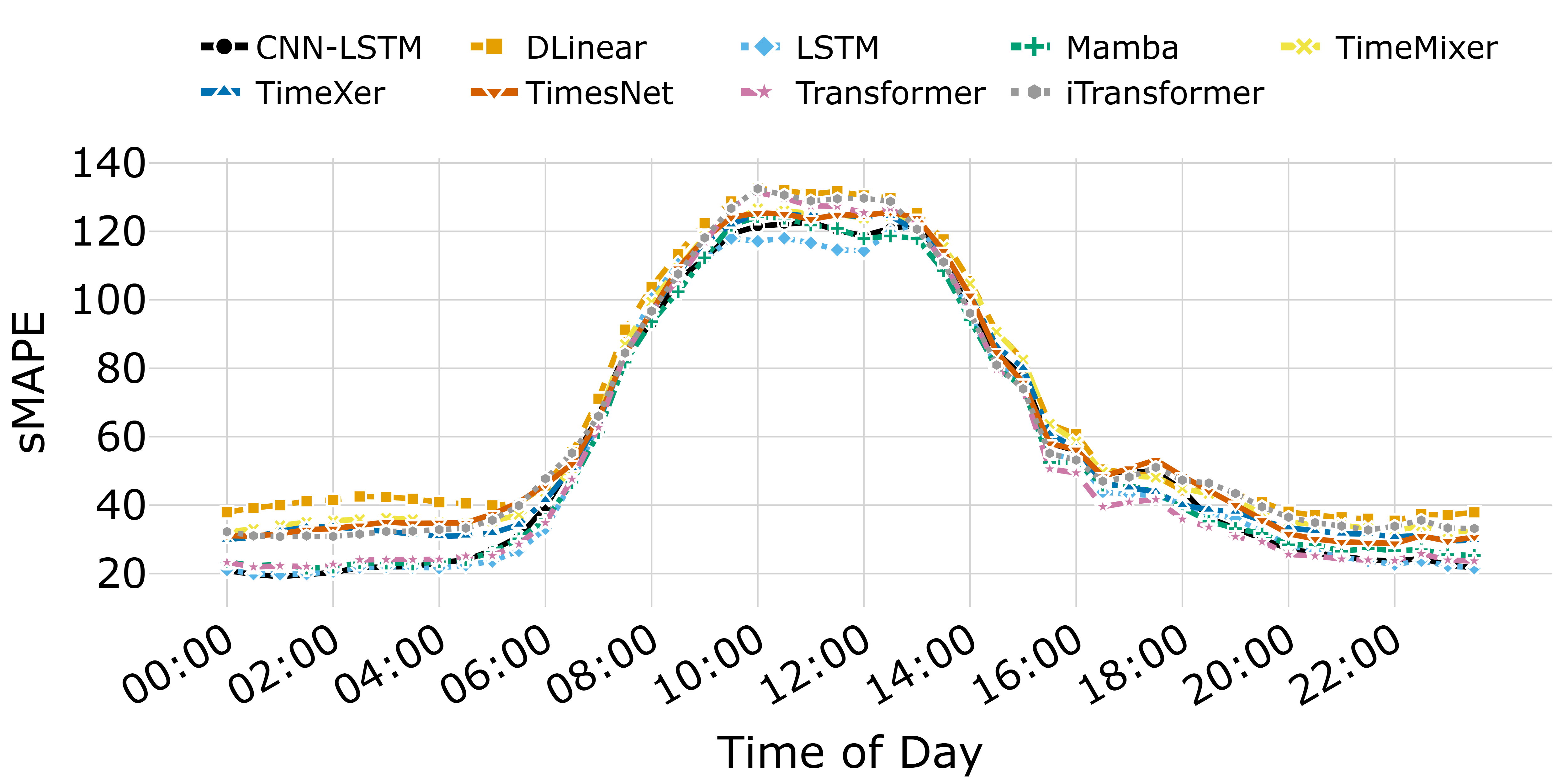}%
  \label{fig:qld_sMAPE}%
}



\subfloat[Diurnal electricity price patterns in the QLD region. The line represents the average price per 30-minute interval across all days, while the bars indicate the percentage of negative prices at each interval. The shaded region (07:00–16:00) highlights the hours with higher sMAPE.]{%
  \includegraphics[width=\linewidth]{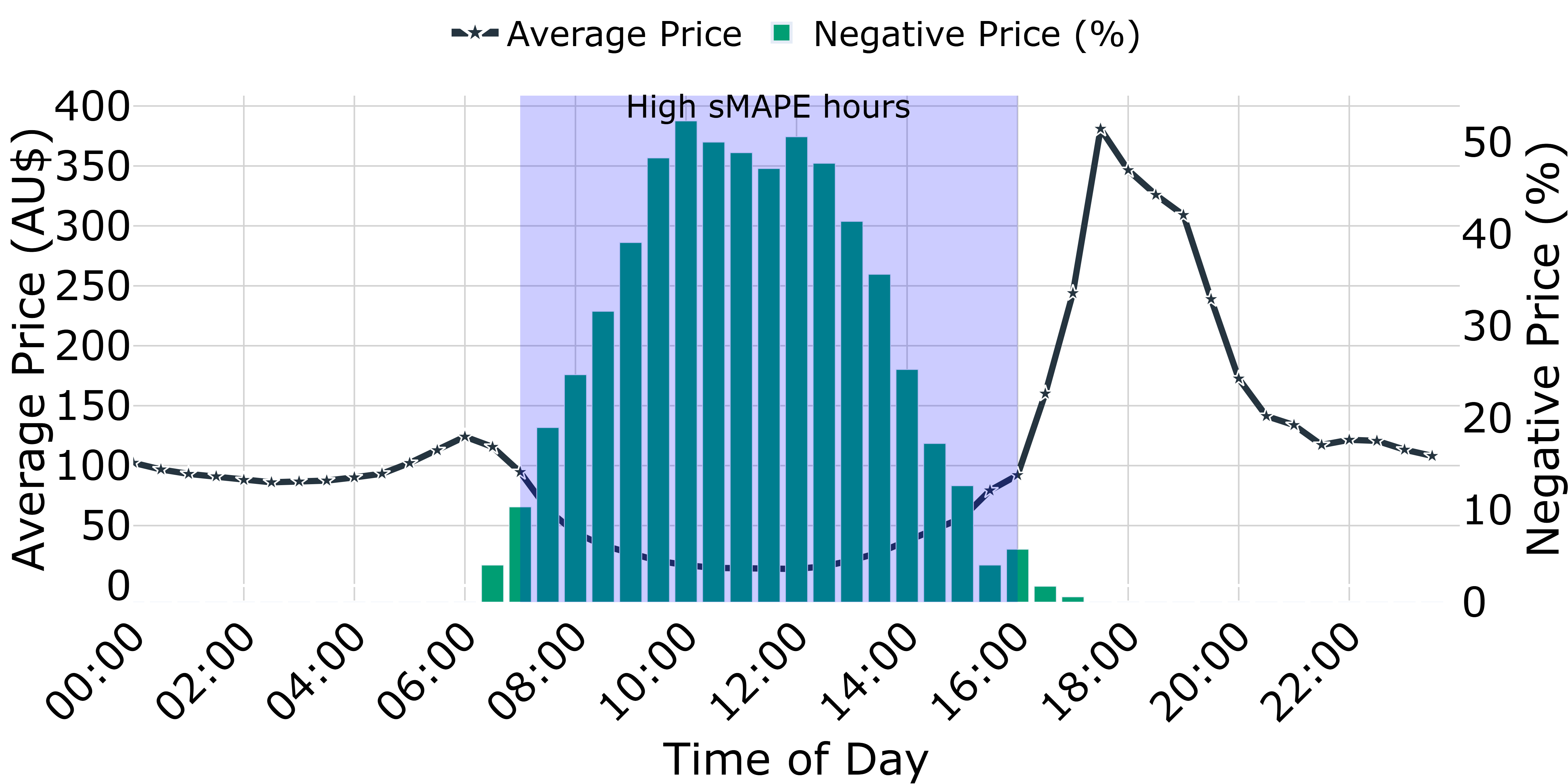}%
  \label{fig:qld_sMAPE_hours}%
}

\caption{Intraday interval-level sMAPE for the \textbf{QLD} region, shown alongside average intraday prices and the prevalence of negative prices.}
\label{fig:qld_sMAPE_combined}
\end{figure}

\subsubsection{Relative Errors Escalate Under Midday Negative‑Price Regimes}\label{negative_price}
Relative error, as measured by sMAPE (Figure~\ref{fig:qld_sMAPE}), reveals a clear intra‑day structure in forecast errors, with performance strongly influenced by the underlying price regime. Figure~\ref{fig:qld_sMAPE_hours} overlays the average price (blue line) with the percentage of negative prices (bars) across intervals, highlighting ``high sMAPE hours'' in light blue where errors are most pronounced. sMAPE spikes dramatically for all models during midday intervals (roughly 07:00–16:00), coinciding with peak negative price occurrences exceeding 40\% in some slots, a consequence of QLD's solar oversupply. When solar generation floods the market, prices often fall below zero to incentivise consumption or curtailment.
Outside these hours, sMAPE stabilises, particularly during evening peaks when average prices rise above 200 A\$/MWh and negative incidences drop to near zero. 
Accordingly, the results indicate that sMAPE degradation is largely driven by negative price intervals, whose prevalence increases with high renewable penetration. This helps explain why regions like SA, with even higher renewable shares (over 75\%)\footnote{\url{https://explore.openelectricity.org.au/energy/sa1/?range=28d&interval=30m&view=time-of-day&group=Detailed}}, face elevated sMAPE overall (Table~\ref{tab:all_metrics}). 

\begin{figure}[!t]
\centering


\subfloat[MDA of all models in the QLD region, evaluated at 30-minute intervals.]{%
  \includegraphics[width=\linewidth]{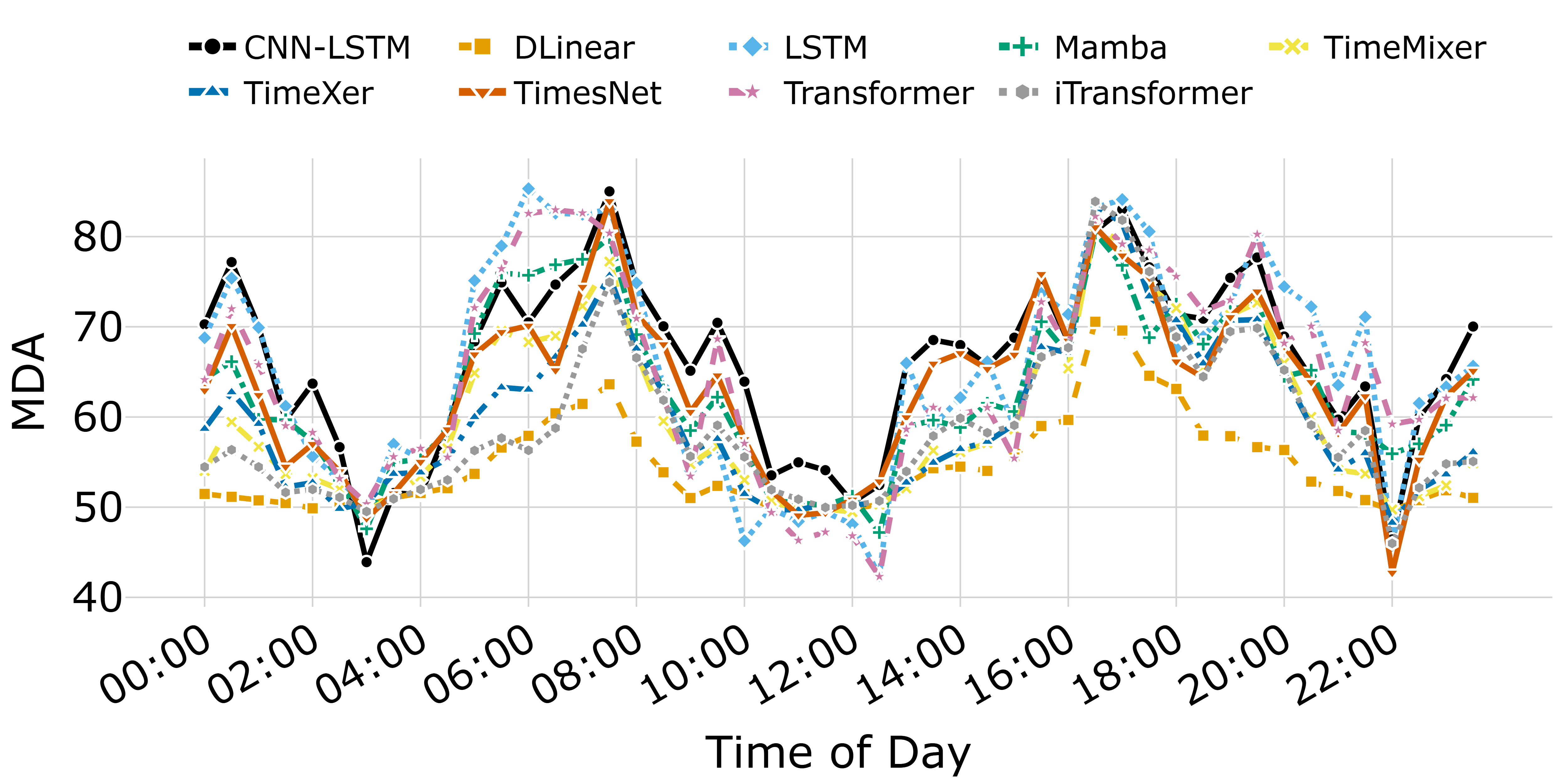}%
  \label{fig:qld_MDA}%
}


\subfloat[The percentage of directional shifts for each interval in the QLD region. Shaded regions indicate the hours with the low MDA, highlighting periods of reduced directional accuracy associated with rapid directional changes.]{%
  \includegraphics[width=\linewidth]{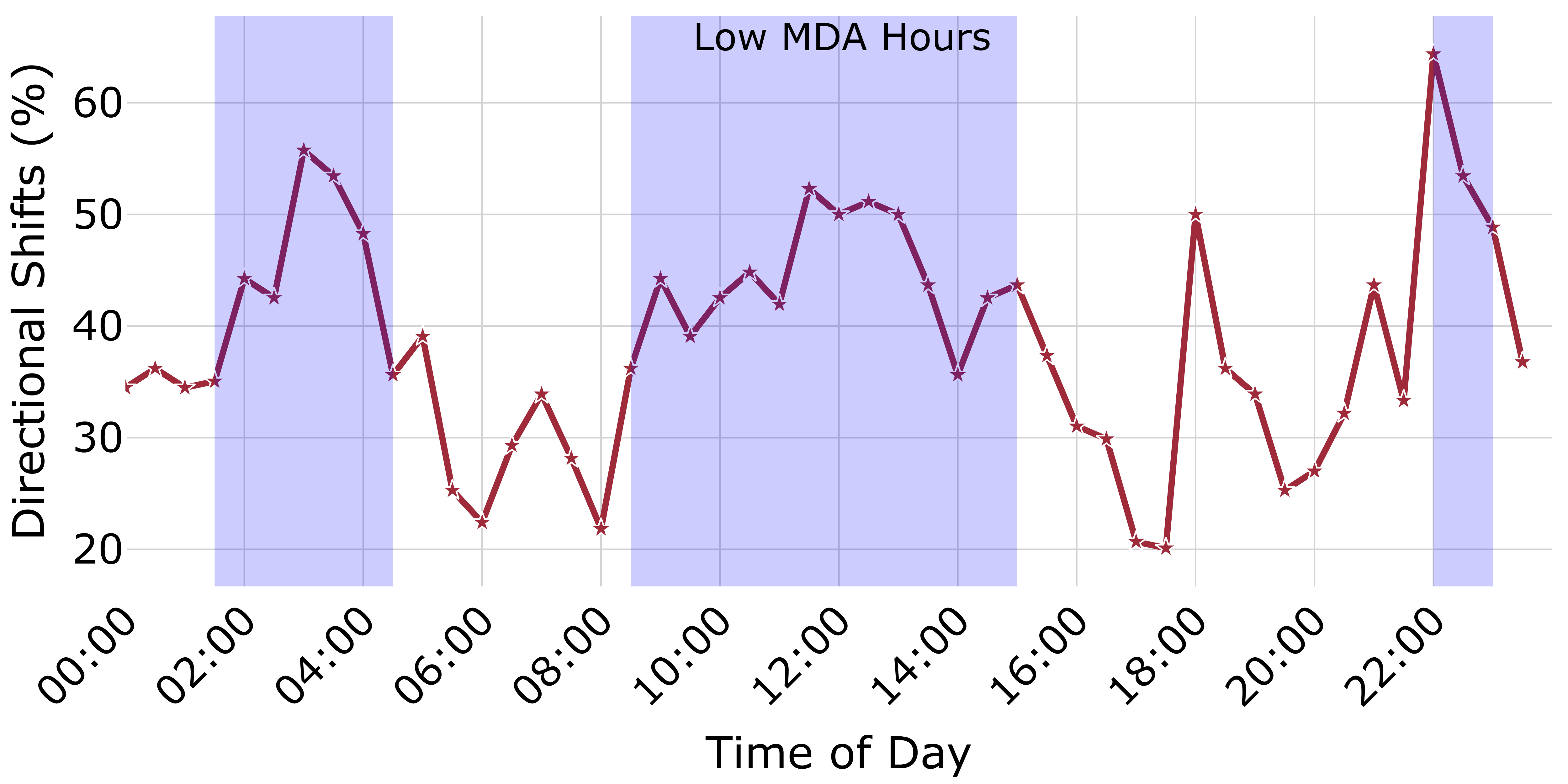}%
  \label{fig:qld_MDA_hours}%
}

\caption{Intraday interval-level MDA for the \textbf{QLD} region and corresponding intraday directional-shift percentage.} 
\label{fig:qld_MDA_combined}
\end{figure}

\subsubsection{Directional Accuracy Degradation during Periods of Frequent Price Shifts}
Directional accuracy, measured through MDA (Figure~\ref{fig:qld_MDA}), shows pronounced intra‑day fluctuations for all models, with noticeable drops during intervals marked by frequent price direction changes. Figure~\ref{fig:qld_MDA_hours} illustrates the percentage of directional shifts in the actual data, enabling the identification of hours during which directional forecasting becomes more challenging.

The first set of periods—01:30–04:30, 08:30–15:00, and 22:00–23:00—exhibit unstable price trends. During these windows, directional shifts frequently exceed 40\%, resulting in the lowest MDA values (shaded region). These intervals are characterised by irregular price movements and rapid direction flips, limiting the ability of forecasting models to maintain sustained directional accuracy. The second set of periods—05:00–08:00 and 15:30–21:30—shows comparatively stable behaviour. Directional shifts during these hours often fall below 40\%, and MDA rises accordingly, frequently exceeding 70\%. Price movements in these intervals follow more consistent upward or downward trajectories, allowing forecasts to align more reliably with the prevailing trend. These intra-day fluctuations are observed across all models, yet DLinear displays persistently lower MDA—even in periods when other models attain elevated directional accuracy.

\subsection{Discussion}
This study shows that, across most NEM regions and for both 24‑ and 48‑hour horizons, standard DL models—LSTM, CNN‑LSTM, and Transformer—generally outperform many recently proposed SOTA time‑series models. While the baselines are not uniformly superior, they remain consistently strong and often top‑ranked, with only a few exceptions—most notably TimeXer, which challenges the baselines in VIC and at the 48‑hour horizon in a few other instances, and Mamba, which is competitive overall. These performance patterns persist even under the most challenging regimes: in the upper and lower 5\% price extremes, TimeXer achieves the lowest errors, but LSTM remains highly competitive and generally surpasses most SOTA models; and during negative‑price intervals, baselines such as LSTM and CNN‑LSTM again outperform nearly all recent models, despite the instability that drives high sMAPE across all models. In contrast, DLinear, TimeMixer, TimesNet, and iTransformer tend to underperform both under typical conditions and in these stress regime scenarios. 

The diurnal interval‑level analysis clarifies why many SOTA time-series models underperform in this setting. Absolute errors spike during the evening volatility window as solar output collapses; sMAPE inflates at midday when zero/negative prices become prevalent; and MDA deteriorates in intervals with frequent trend change. Across these structural regime shifts, the SOTA time‑series models degrade more sharply, indicating poorer adaptation precisely when market conditions change abruptly. These patterns suggest that many recent SOTA architectures—often tuned to datasets with stable periodicity, limited volatility, and few structural breaks—generalise poorly to the NEM’s non‑stationary, regime‑switching reality. More broadly, these findings highlight limitations in prevailing time-series benchmarking practices, which often under-represent volatility, structural breaks, and negative pricing. From an operational perspective, the results suggest that well-tuned standard DL models may currently offer more reliable performance for EPF than newer architectures despite their theoretical appeal.

The findings motivate several targeted directions for future research. 
\emph{First}, extending forecasting frameworks to multivariate settings with effective feature representations—including demand, renewable output, interconnector flows, and other system variables—may allow models to better contextualise regime transitions.
\emph{Second}, developing volatility‑aware or regime‑adaptive architectures, such as models that explicitly detect spikes, integrate latent regime‑switching dynamics, or modulate their forecasting mechanism based on price conditions, may better handle the irregular behaviour of electricity prices. 
\emph{Third}, designing loss functions tailored to electricity‑market realities—for example, asymmetric penalties for spikes, relative‑error formulations that correctly handle negative prices, or hybrid objectives that simultaneously optimise price‑level accuracy, directional prediction, and robustness to volatility fluctuations—could enhance model learning during periods of abrupt regime transitions. Together, these directions point toward forecasting approaches that better reflect the realities of renewable‑rich, volatility‑prone electricity systems.

\section{Conclusion}\label{sec5}
This paper has provided a systematic evaluation of SOTA deep time-series models for multi-horizon EPF in the highly volatile NEM. Across many regions and forecast horizons, the empirical results show that standard DL baselines frequently outperform recent SOTA time-series models, including under extreme-price and negative-price conditions. The findings underscore the gap between prevailing time‑series benchmarking expectations and real‑world electricity markets, indicating that these models do not readily transfer to a volatility‑prone market. The intraday analysis further demonstrates that forecasting accuracy deteriorates most during abrupt market transitions, underscoring the difficulty these models face in adapting to rapidly changing dynamics. Overall, the findings suggest that advancing EPF in volatility‑prone markets will require models that explicitly account for regime‑switching behaviour and incorporate richer contextual features reflective of underlying system conditions.

\bibliographystyle{IEEEtran}
\bibliography{references}

@ARTICLE{Wang2024-ky,
  title         = "Deep time series models: A comprehensive survey and benchmark",
  author        = "Wang, Yuxuan and Wu, Haixu and Dong, Jiaxiang and Liu, Yong
                   and Long, Mingsheng and Wang, Jianmin",
  journal       = "ArXiv",
  year          =  2024,
  archivePrefix = "arXiv",
  primaryClass  = "cs.LG",
  doi           = "10.48550/arXiv.2407.13278"
}

@INPROCEEDINGS{Gu2024-ve,
  title     = "Mamba: Linear-Time Sequence Modeling with Selective State Spaces",
  author    = "Gu, Albert and Dao, Tri",
  booktitle = "First Conference on Language Modeling",
  year      =  2024,
  doi       = "10.48550/arXiv.2312.00752"
}

@ARTICLE{Chai2024-gi,
  title     = "Forecasting electricity prices from the state-of-the-art modeling
               technology and the price determinant perspectives",
  author    = "Chai, Shanglei and Li, Qiang and Abedin, Mohammad Zoynul and
               Lucey, Brian M",
  journal   = "Res. Int. Bus. Fin.",
  publisher = "Elsevier BV",
  volume    =  67,
  pages     =  102132,
  year      =  2024,
  doi       = "10.1016/j.ribaf.2023.102132"
}

@ARTICLE{Lago2021-oj,
  title     = "Forecasting day-ahead electricity prices: A review of
               state-of-the-art algorithms, best practices and an open-access
               benchmark",
  author    = "Lago, Jesus and Marcjasz, Grzegorz and De Schutter, Bart and
               Weron, Rafał",
  journal   = "Appl. Energy",
  publisher = "Elsevier BV",
  volume    =  293,
  pages     =  116983,
  year      =  2021,
  doi       = "10.1016/j.apenergy.2021.116983"
}

@ARTICLE{Wang2025-mt,
  title     = "{TSGT}: Electricity price prediction model based on
               Time-Space-{GCN}-Transformer",
  author    = "Wang, Feng and Zhu, Guangping and Pu, Tong and Xiang, Dong",
  journal   = "Procedia Comput. Sci.",
  publisher = "Elsevier BV",
  volume    =  266,
  pages     = "357--364",
  year      =  2025,
  doi       = "10.1016/j.procs.2025.08.045"
}

@ARTICLE{Tan2023-ck,
  title     = "Day-ahead electricity price forecasting employing a novel hybrid
               frame of deep learning methods: A case study in {NSW}, Australia",
  author    = "Tan, Yong Qiang and Shen, Yan Xia and Yu, Xin Yan and Lu, Xin",
  journal   = "Electric Power Syst. Res.",
  publisher = "Elsevier BV",
  volume    =  220,
  pages     =  109300,
  year      =  2023,
  doi       = "10.1016/j.epsr.2023.109300"
}

@ARTICLE{Lago2018-tc,
  title     = "Forecasting spot electricity prices: Deep learning approaches and
               empirical comparison of traditional algorithms",
  author    = "Lago, Jesus and De Ridder, Fjo and De Schutter, Bart",
  journal   = "Appl. Energy",
  publisher = "Elsevier BV",
  volume    =  221,
  pages     = "386--405",
  year      =  2018,
  doi       = "10.1016/j.apenergy.2018.02.069"
}

@ARTICLE{Mubarak2024-rs,
  title     = "Day-Ahead electricity price forecasting using a {CNN}-{BiLSTM}
               model in conjunction with autoregressive modeling and
               hyperparameter optimization",
  author    = "Mubarak, Hamza and Abdellatif, Abdallah and Ahmad, Shameem and
               Zohurul Islam, Mohammad and Muyeen, S M and Abdul Mannan,
               Mohammad and Kamwa, Innocent",
  journal   = "Int. J. Electr. Power Energy Syst.",
  publisher = "Elsevier BV",
  volume    =  161,
  pages     =  110206,
  year      =  2024,
  doi       = "10.1016/j.ijepes.2024.110206"
}

@ARTICLE{Li2024-yh,
  title     = "Joint forecasting of source-load-price for integrated energy
               system based on multi-task learning and hybrid attention
               mechanism",
  author    = "Li, Ke and Mu, Yuchen and Yang, Fan and Wang, Haiyang and Yan, Yi
               and Zhang, Chenghui",
  journal   = "Appl. Energy",
  publisher = "Elsevier BV",
  volume    =  360,
  pages     =  122821,
  year      =  2024,
  doi       = "10.1016/j.apenergy.2024.122821"
}

@ARTICLE{Prakash2023-to,
  title     = "Electricity price forecasting using hybrid deep learned networks",
  author    = "Prakash, Krishna and Singh, Jai Govind",
  journal   = "J. Forecast.",
  publisher = "Wiley",
  volume    =  42,
  number    =  7,
  pages     = "1750--1771",
  year      =  2023,
  doi       = "10.1002/for.2981"
}

@ARTICLE{Huang2021-zb,
  title     = "A novel hybrid deep neural network model for short‐term
               electricity price forecasting",
  author    = "Huang, Chiou-Jye and Shen, Yamin and Chen, Yung-Hsiang and Chen,
               Hsin-Chuan",
  journal   = "Int. J. Energy Res.",
  publisher = "Hindawi Limited",
  volume    =  45,
  number    =  2,
  pages     = "2511--2532",
  year      =  2021,
  doi       = "10.1002/er.5945"
}

@ARTICLE{Kuo2018-dk,
  title     = "An electricity price forecasting model by hybrid structured deep
               neural networks",
  author    = "Kuo, Ping-Huan and Huang, Chiou-Jye",
  journal   = "Sustainability",
  publisher = "MDPI AG",
  volume    =  10,
  number    =  4,
  pages     =  1280,
  year      =  2018,
  doi       = "10.3390/su10041280"
}

@ARTICLE{Li2021-bq,
  title     = "Day-ahead electricity price prediction applying hybrid models of
               {LSTM}-based deep learning methods and feature selection
               algorithms under consideration of market coupling",
  author    = "Li, Wei and Becker, Denis Mike",
  journal   = "Energy",
  publisher = "Elsevier BV",
  volume    =  237,
  pages     =  121543,
  year      =  2021,
  doi       = "10.1016/j.energy.2021.121543"
}

@ARTICLE{Li2022-nq,
  title     = "Dense skip attention based deep learning for day-ahead
               electricity price forecasting",
  author    = "Li, Yuanzheng and Ding, Yizhou and Liu, Yun and Yang, Tao and
               Wang, Ping and Wang, Jingfei and Yao, Wei",
  journal   = "IEEE Trans. Power Syst.",
  publisher = "Institute of Electrical and Electronics Engineers (IEEE)",
  volume    =  38,
  number    =  5,
  pages     = "4308--4327",
  year      =  2022,
  doi       = "10.1109/tpwrs.2022.3217579"
}

@ARTICLE{Khan2025-mj,
  title     = "Enhanced transformer-{BiLSTM} deep learning framework for
               day-ahead energy price forecasting",
  author    = "Khan, Abdullah Al Ahad and Ullah, Md Habib and Tabassum, Ruchira
               and Kabir, Md Faisal",
  journal   = "IEEE Trans. Ind. Appl.",
  publisher = "Institute of Electrical and Electronics Engineers (IEEE)",
  pages     = "1--15",
  year      =  2025,
  doi       = "10.1109/tia.2025.3599812"
}

@ARTICLE{Heidarpanah2023-ax,
  title     = "Daily electricity price forecasting using artificial intelligence
               models in the Iranian electricity market",
  author    = "Heidarpanah, Mohammadreza and Hooshyaripor, Farhad and Fazeli,
               Meysam",
  journal   = "Energy",
  publisher = "Elsevier BV",
  volume    =  263,
  pages     =  126011,
  year      =  2023,
  doi       = "10.1016/j.energy.2022.126011"
}

@ARTICLE{Ehsani2024-cz,
  title     = "Price forecasting in the Ontario electricity market via
               {TriConvGRU} hybrid model: Univariate vs. multivariate frameworks",
  author    = "Ehsani, Behdad and Pineau, Pierre-Olivier and Charlin, Laurent",
  journal   = "Appl. Energy",
  publisher = "Elsevier BV",
  volume    =  359,
  pages     =  122649,
  year      =  2024,
  doi       = "10.1016/j.apenergy.2024.122649"
}

@ARTICLE{Ghimire2024-ou,
  title     = "Two-step deep learning framework with error compensation
               technique for short-term, half-hourly electricity price
               forecasting",
  author    = "Ghimire, Sujan and Deo, Ravinesh C and Casillas-Pérez, David and
               Salcedo-Sanz, Sancho",
  journal   = "Appl. Energy",
  publisher = "Elsevier BV",
  volume    =  353,
  pages     =  122059,
  year      =  2024,
  doi       = "10.1016/j.apenergy.2023.122059"
}

@ARTICLE{Yang2022-wu,
  title     = "Short‐term electricity price forecasting based on graph
               convolution network and attention mechanism",
  author    = "Yang, Yuyun and Tan, Zhenfei and Yang, Haitao and Ruan, Guangchun
               and Zhong, Haiwang and Liu, Fengkui",
  journal   = "IET Renew. Power Gener.",
  publisher = "Institution of Engineering and Technology (IET)",
  volume    =  16,
  number    =  12,
  pages     = "2481--2492",
  year      =  2022,
  doi       = "10.1049/rpg2.12413"
}

@ARTICLE{Pourdaryaei2024-kb,
  title     = "A new framework for electricity price forecasting via multi-head
               self-attention and {CNN}-based techniques in the competitive
               electricity market",
  author    = "Pourdaryaei, Alireza and Mohammadi, Mohammad and Mubarak, Hamza
               and Abdellatif, Abdallah and Karimi, Mazaher and Gryazina, Elena
               and Terzija, Vladimir",
  journal   = "Expert Syst. Appl.",
  publisher = "Elsevier BV",
  volume    =  235,
  pages     =  121207,
  year      =  2024,
  doi       = "10.1016/j.eswa.2023.121207"
}

@ARTICLE{Chang2019-gg,
  title     = "Electricity price prediction based on hybrid model of adam
               optimized {LSTM} neural network and wavelet transform",
  author    = "Chang, Zihan and Zhang, Yang and Chen, Wenbo",
  journal   = "Energy",
  publisher = "Elsevier BV",
  volume    =  187,
  pages     =  115804,
  year      =  2019,
  doi       = "10.1016/j.energy.2019.07.134"
}

@INPROCEEDINGS{Nie2023-lh,
  title     = "A Time Series is Worth 64 Words: Long-term Forecasting with
               Transformers",
  author    = "Nie, Yuqi and Nguyen, Nam H and Sinthong, Phanwadee and
               Kalagnanam, J",
  booktitle = "International Conference on Learning Representations (ICLR)",
  year      =  2023,
  doi       = "10.48550/arXiv.2211.14730"
}

@INPROCEEDINGS{Zhou2021-yd,
  title     = "Informer: Beyond efficient transformer for long sequence
               time-series forecasting",
  author    = "Zhou, Haoyi and Zhang, Shanghang and Peng, Jieqi and Zhang, Shuai
               and Li, Jianxin and Xiong, Hui and Zhang, Wancai",
  booktitle = "Proceedings of the Thirty-Fifth AAAI Conference on Artificial
               Intelligence",
  publisher = "Association for the Advancement of Artificial Intelligence (AAAI)",
  volume    =  35,
  pages     = "11106--11115",
  year      =  2021,
  doi       = "10.1609/aaai.v35i12.17325"
}

@INPROCEEDINGS{Oreshkin2020-pe,
  title     = "{N}-{BEATS}: Neural basis expansion analysis for interpretable
               time series forecasting",
  author    = "Oreshkin, Boris N and Carpov, Dmitri and Chapados, Nicolas and
               Bengio, Yoshua",
  booktitle = "International Conference on Learning Representations (ICLR)",
  year      =  2020,
  doi       = "10.48550/arXiv.1905.10437"
}

@INPROCEEDINGS{Liu2024-sl,
  title     = "{iTransformer}: Inverted Transformers are effective for time
               series forecasting",
  author    = "Liu, Yong and Hu, Tengge and Zhang, Haoran and Wu, Haixu and
               Wang, Shiyu and Ma, Lintao and Long, Mingsheng",
  booktitle = "International Conference on Learning Representations (ICLR)",
  year      =  2024,
  doi       = "10.48550/arXiv.2310.06625"
}

@INPROCEEDINGS{Wu2023-cm,
  title     = "{TimesNet}: Temporal {2D}-variation modeling for general time
               series analysis",
  author    = "Wu, Haixu and Hu, Teng and Liu, Yong and Zhou, Hang and Wang,
               Jianmin and Long, Mingsheng",
  booktitle = "International Conference on Learning Representations (ICLR)",
  year      =  2023,
  doi       = "10.48550/arXiv.2210.02186"
}

@INPROCEEDINGS{Liu2022-ar,
  title     = "{SCINet}: Time series modeling and forecasting with sample
               convolution and interaction",
  author    = "Liu, Minhao and Zeng, Ailing and Chen, Mu-Hwa and Xu, Zhijian and
               Lai, Qiuxia and Ma, Lingna and Xu, Qiang",
  booktitle = "Advances in Neural Information Processing Systems",
  publisher = "Curran Associates, Inc.",
  volume    =  35,
  pages     = "5816--5828",
  year      =  2022,
  doi       = "10.48550/arXiv.2106.09305"
}

@INPROCEEDINGS{Zhou2022-vq,
  title     = "{FEDformer}: Frequency Enhanced Decomposed Transformer for
               long-term series forecasting",
  author    = "Zhou, Tian and Ma, Ziqing and Wen, Qingsong and Wang, Xue and
               Sun, Liang and Jin, Rong",
  booktitle = "Proceedings of the 39th International Conference on Machine
               Learning",
  publisher = "PMLR",
  volume    =  162,
  pages     = "27268--27286",
  year      =  2022,
  doi       = "10.48550/arXiv.2201.12740"
}

@INPROCEEDINGS{Gu2022-yb,
  title     = "Efficiently modeling long sequences with structured state spaces",
  author    = "Gu, Albert and Goel, Karan and R'e, Christopher",
  booktitle = "International Conference on Learning Representations (ICLR)",
  year      =  2022,
  doi       = "10.48550/arXiv.2111.00396"
}

@INPROCEEDINGS{Wu2021-sp,
  title     = "Autoformer: Decomposition Transformers with Auto-Correlation for
               long-term series forecasting",
  author    = "Wu, Haixu and Xu, Jiehui and Wang, Jianmin and Long, Mingsheng",
  booktitle = "Advances in Neural Information Processing Systems",
  publisher = "Curran Associates, Inc.",
  volume    =  34,
  pages     = "22419--22430",
  year      =  2021,
  doi       = "10.48550/arXiv.2106.13008"
}

@INPROCEEDINGS{Zeng2023-zt,
  title     = "Are Transformers effective for time series forecasting?",
  author    = "Zeng, Ailing and Chen, Muxi and Zhang, Lei and Xu, Qiang",
  booktitle = "Proceedings of the Thirty-Seventh AAAI Conference on Artificial
               Intelligence",
  publisher = "Association for the Advancement of Artificial Intelligence (AAAI)",
  volume    =  37,
  pages     = "11121--11128",
  year      =  2023,
  doi       = "10.1609/aaai.v37i9.26317"
}

@INPROCEEDINGS{Wang2024-eg,
  title     = "{TimeMixer}: Decomposable Multiscale Mixing for Time Series
               Forecasting",
  author    = "Wang, Shiyu and Wu, Haixu and Shi, Xiaoming and Hu, Tengge and
               Luo, Huakun and Ma, Lintao and Zhang, James Y and Zhou, Jun",
  booktitle = "International Conference on Learning Representations (ICLR)",
  year      =  2024,
  doi       = "10.48550/arXiv.2405.14616"
}

@INPROCEEDINGS{Wang2024-iy,
  title     = "{TimeXer}: Empowering Transformers for Time Series Forecasting
               with Exogenous Variables",
  author    = "Wang, Yuxuan and Wu, Haixu and Dong, Jiaxiang and Qin, Guo and
               Zhang, Haoran and Liu, Yong and Qiu, Yunzhong and Wang, Jianmin
               and Long, Mingsheng",
  booktitle = "Advances in Neural Information Processing Systems",
  pages     = "469–498",
  year      =  2024,
  doi       = "10.48550/arXiv.2402.19072"
}

@ARTICLE{Keles2016-tr,
  title     = "Extended forecast methods for day-ahead electricity spot prices
               applying artificial neural networks",
  author    = "Keles, Dogan and Scelle, Jonathan and Paraschiv, Florentina and
               Fichtner, Wolf",
  journal   = "Appl. Energy",
  publisher = "Elsevier BV",
  volume    =  162,
  pages     = "218--230",
  year      =  2016,
  doi       = "10.1016/j.apenergy.2015.09.087"
}

@ARTICLE{Lago2018-sh,
  title     = "Forecasting day-ahead electricity prices in Europe: The
               importance of considering market integration",
  author    = "Lago, Jesus and De Ridder, Fjo and Vrancx, Peter and De Schutter,
               Bart",
  journal   = "Appl. Energy",
  publisher = "Elsevier BV",
  volume    =  211,
  pages     = "890--903",
  year      =  2018,
  doi       = "10.1016/j.apenergy.2017.11.098"
}

@ARTICLE{Perla2025-he,
  title     = "Short-term forecasting of electricity price using ensemble deep
               kernel based random vector functional link network",
  author    = "Perla, Someswari and Bisoi, Ranjeeta and Dash, P K and Rout, A K",
  journal   = "Appl. Soft Comput.",
  publisher = "Elsevier BV",
  volume    =  174,
  pages     =  113012,
  year      =  2025,
  doi       = "10.1016/j.asoc.2025.113012"
}

@ARTICLE{Li2025-zw,
  title     = "Short-term electricity price forecasting via {CPO}-enhanced dual
               decomposition and {NRBO}-optimized deep learning",
  author    = "Li, Muan and Wang, Xiao",
  journal   = "Digit. Signal Process.",
  publisher = "Elsevier BV",
  volume    =  168,
  pages     =  105520,
  year      =  2025,
  doi       = "10.1016/j.dsp.2025.105520"
}

@ARTICLE{Deng2022-qo,
  title     = "Operational scheduling of behind-the-meter storage systems based
               on multiple nonstationary decomposition and deep convolutional
               neural network for price forecasting",
  author    = "Deng, Zhuofu and Qi, Xianglong and Xu, Tengteng and Zheng,
               Yingnan",
  journal   = "Comput. Intell. Neurosci.",
  publisher = "Hindawi Limited",
  volume    =  2022,
  number    =  1,
  pages     =  9326856,
  year      =  2022,
  doi       = "10.1155/2022/9326856"
}

@ARTICLE{Iwabuchi2022-ou,
  title     = "Flexible electricity price forecasting by switching mother
               wavelets based on wavelet transform and Long Short-Term Memory",
  author    = "Iwabuchi, Koki and Kato, Kenshiro and Watari, Daichi and
               Taniguchi, Ittetsu and Catthoor, Francky and Shirazi, Elham and
               Onoye, Takao",
  journal   = "Energy and AI",
  publisher = "Elsevier BV",
  volume    =  10,
  pages     =  100192,
  year      =  2022,
  doi       = "10.1016/j.egyai.2022.100192"
}

@ARTICLE{Numminen2025-wf,
  title     = "Shifting toward dynamic pricing of electricity: What did we learn
               from the {2021–2024} energy crises?",
  author    = "Numminen, Sini and Jalas, Mikko and Ruggiero, Salvatore and Värä,
               Arina",
  journal   = "Smart Energy",
  publisher = "Elsevier BV",
  volume    =  20,
  number    =  100210,
  pages     =  100210,
  year      =  2025,
  doi       = "10.1016/j.segy.2025.100210"
}


 





\end{document}